\documentclass[twocolumn, DIV22, 9pt]{scrartcl}
\pdfoutput=1 
\usepackage{tgschola}
\usepackage{helvet}
\usepackage[inline=true,margin=false,draft]{fixme}
\usepackage[pdftex,
colorlinks=true,
linkcolor=blue,
citecolor=black,
urlcolor=blue,
hyperfootnotes=false]{hyperref}
\usepackage{stfloats}
\fxusetheme{color}
\usepackage{graphicx} 
\usepackage{color}
\usepackage{subcaption}
\usepackage{url}
\usepackage[backend=biber,sorting=none,url=false,style=nature]{biblatex}
\hypersetup{hidelinks}
\renewcommand{\cite}[1]{\supercite{#1}}

\DeclareFieldFormat{labelnumberwidth}{\mkbibparens{#1}}
\DeclareFieldFormat{shorthandwidth}{\mkbibparens{#1}}

\usepackage{xcolor}

\definecolor{blue}{rgb}{0.31,0.44,0.66}

\usepackage[normalem]{ulem}
\usepackage{todonotes}
\usepackage{amsmath}
\usepackage{amsmath,bm}
\usepackage{amssymb}
\usepackage{multicol}
\usepackage{booktabs}

\usepackage[labelfont={bf,small,sf}]{caption}
\DeclareCaptionLabelSeparator{pp}{.}
\setcapindent{0cm}
\DeclareCaptionType{si-figure}[Fig.]
\DeclareCaptionLabelFormat{bf-parens}{#1 S#2.}
\usepackage{fancyhdr}
\newcommand{\raisedrule}[2][0em]{\leaders\hbox{\rule[#1]{15pt}{#2}}\hfill}
\makeatletter
\newcommand{\removelatexerror}{\let\@latex@error\@gobble}
\makeatother
\let\oldnl\nl
\newcommand{\nonl}{\renewcommand{\nl}{\let\nl\oldnl}}

\date{}

\begin{document}
\pagestyle{fancy}
\makeatletter
\twocolumn[
   \begin{@twocolumnfalse} 
    {\fontsize{25}{12}\selectfont
    \textsf{\textbf{Flying in air ducts}}}
    
    \bigskip
    \sffamily
    \textbf{Thomas Martin$^{1}$, Adrien Guénard$^{2}$, Vladislav Tempez$^{1}$, Lucien Renaud$^{1}$, Thibaut Raharijaona$^{3}$, Franck Ruffier$^{4}$,  Jean-Baptiste Mouret$^{1\ast}$}

    \bigskip   
    
    $^1$Inria, CNRS, Université de Lorraine, LORIA, Nancy, France
    $\bullet$
    $^2$CNRS, Université de Lorraine, LORIA, Nancy, France
    $\bullet$
    $^3$Université de Lorraine, LCFC, Metz, France
    $\bullet$
    $^4$Aix Marseille Univ, CNRS, ISM, Marseille, France
     
     \bigskip
     
     $^\ast$jean-baptiste.mouret@inria.fr
     \bigskip
     
\bfseries
\noindent{}

Air ducts are integral to modern buildings but are challenging to access for inspection. Small quadrotor drones offer a potential solution, as they can navigate both horizontal and vertical sections and smoothly fly over debris. However, hovering inside air ducts is problematic due to the airflow generated by the rotors, which recirculates inside the duct and destabilizes the drone, whereas hovering is a key feature for many inspection missions. In this article, we map the aerodynamic forces that affect a hovering drone in a duct using a robotic setup and a force/torque sensor. Based on the collected aerodynamic data, we identify a recommended position for stable flight, which corresponds to the bottom third for a circular duct. We then develop a neural network-based positioning system that leverages low-cost time-of-flight sensors. By combining these aerodynamic insights and the data-driven positioning system, we show that a small quadrotor drone (here, 180 mm) can hover and fly inside small air ducts, starting with a diameter of 350 mm. These results open a new and promising application domain for drones.

Video: \url{https://youtu.be/BLQqoa7Zolw}

\bigskip
    \end{@twocolumnfalse}
]
\makeatother

\begin{figure*}[!hb]
  \includegraphics*[width=\textwidth]{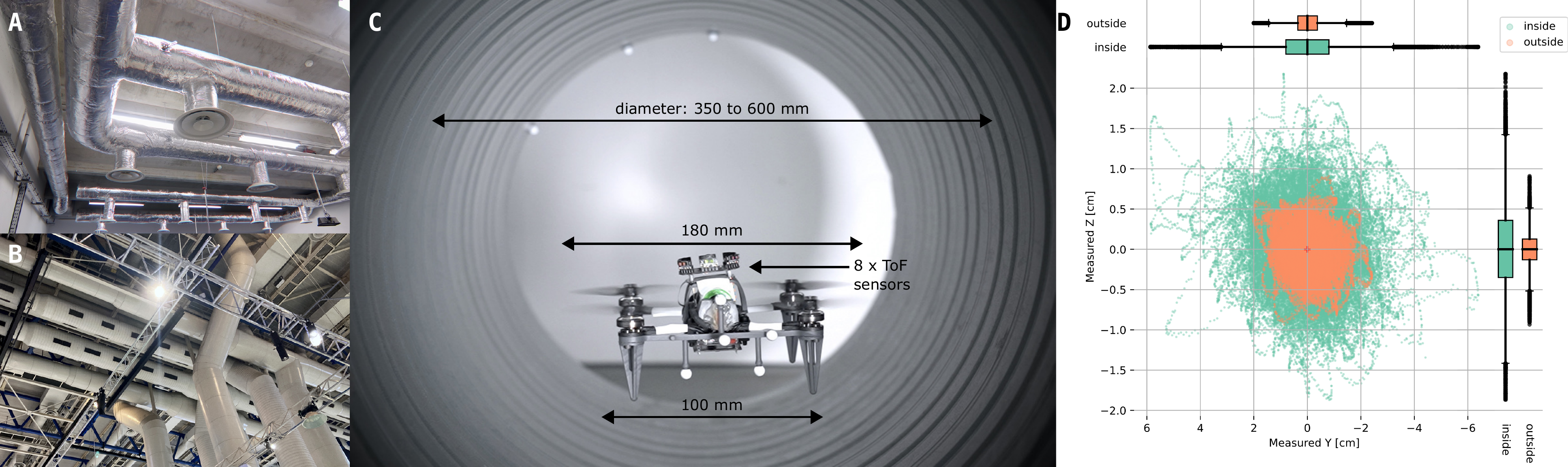}
  \caption{\label{fig:concept}\textbf{A-B. Examples of air duct networks.} Air ducts are common in most modern buildings because they are necessary for maintaining air quality and temperature. \textbf{C. Problem considered} The objective of this article is to make it possible for small quadrotor drones (here, 180mm) to hover (and fly) air ducts with a diameter from 350 to 560 mm. \textbf{D. In an air duct, drones are significantly less stable.} We tested the same drone with the same controller (based on external motion tracking, Methods), inside and outside a circular air duct with a diameter of 35cm cm and tracked the position for 120 seconds. Because of the aerodynamic recirculations inside a duct, the drone has a hard time hovering at a fixed position.}
\end{figure*}

\section*{Introduction}
Air ducts are common sights near the ceiling of most industrial buildings, and are often concealed inside the walls and drop ceilings of offices, hospitals, and modern houses  (Fig.~\ref{fig:concept}.A). They also constitute a significant component in the design of underground networks, such as metro stations. These air ducts are essential for maintaining air quality, heating, and air conditioning. Like most parts of a building, they need regular inspections to detect or identify faults; however, they are inherently impossible to access to humans due to their narrow dimensions and inability to support human weight.

Many robots have been proposed to inspect air ducts \cite{wang2006autonomous,tanise2017development,jeon2012design} and, for a close problem, the numerous sewage, gas and water pipes \cite{roh2005differential,shao2015review}. All of them are based on variations of wheels and tracks and some of them are deployed in industries. Unfortunately, these robots are mostly fit for water pipes or short lengths of air ducts, as they do not cope with the vertical and ascending/descending parts that are common in air duct networks.

In this article, we propose to use quadrotor drones as a new way of accessing and inspecting air ducts. While flying in such a constrained environment might appear counterintuitive, drones can easily go up or down when needed; they can also fly over debris on the ground and are easier to deploy when the entrance of a duct is near the ceiling. Current drones are a mature technology that can be small enough to fit in 35+ cm pipes \cite{giernackiCrazyflieQuadrotorPlatform2017} and can embed mission-specific sensors like gas leak detectors \cite{duisterhof2021sniffy}. Overall, quadrotors have the potential to be small enough and flexible enough to move in complex air ducts and pipes.

Before the 1960s, most air ducts were rectangular because they were easy to manufacture from bended metal sheet and easy to fit in buildings. Since then, engineers realized that round ducts have numerous advantages while being inexpensive to manufacture with modern forming machines. Notably, round ducts offer superior sealing compared to plate assemblies, require less metal for equivalent airflow, and are available in longer lengths. In this study, we examine circular air ducts ranging from 35 to 56 cm in diameter, as these sizes are commonly found in buildings and pose challenges for flying. Ducts smaller than this range are too narrow for 10-20 cm drones, while larger ducts are less common and pose fewer challenges.

\begin{figure*}
  \includegraphics*[width=\textwidth]{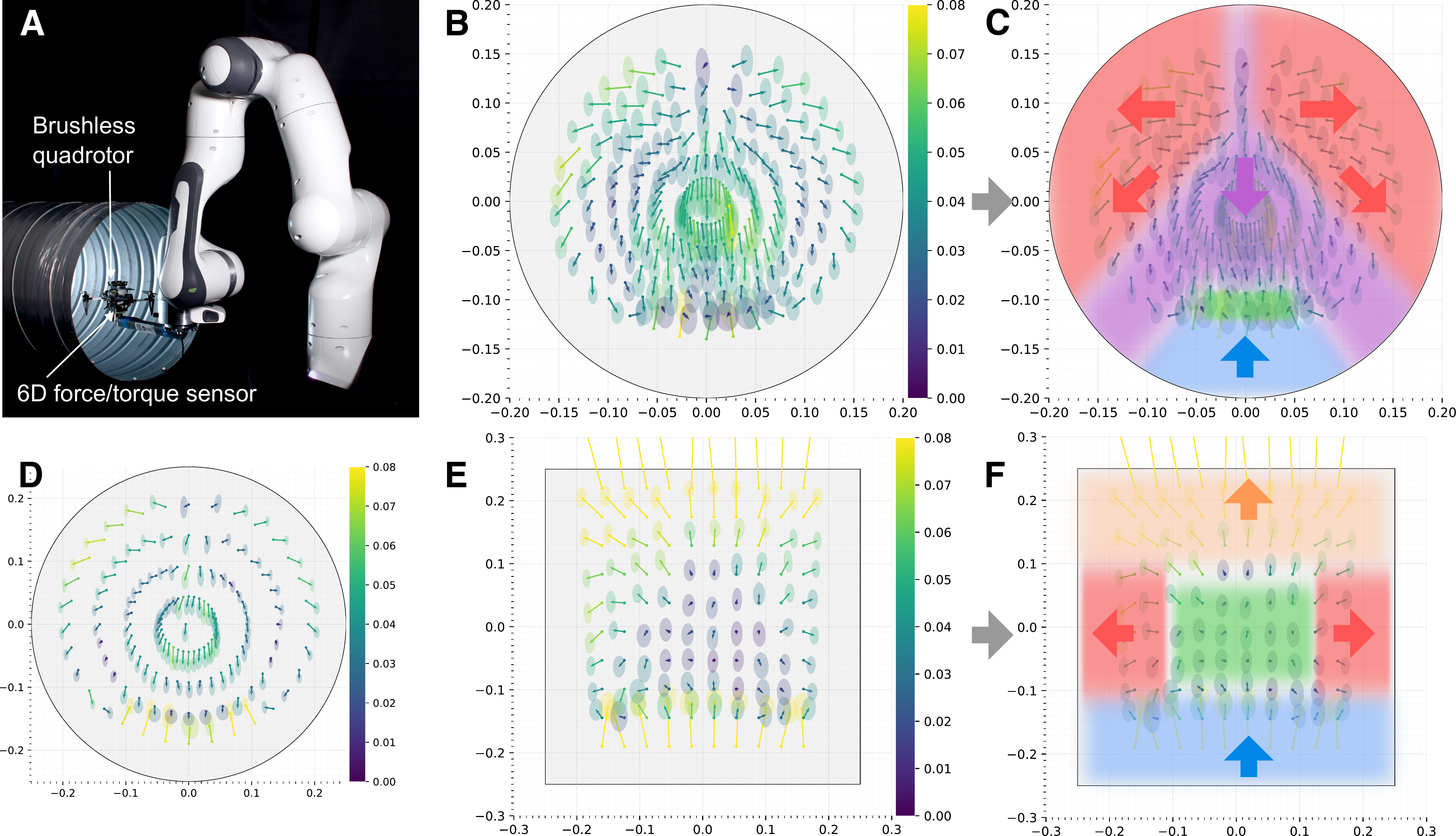}
  \caption{\label{fig:aero}
  \textbf{A. Experimental setup to measure the aerodynamic forces added by the recirculations.} The drone is screwed on a 6-dimensional force/torque sensor, which is fixed to a 7-DOF robotic manipulator. The manipulator makes it possible to measure the forces at 192 different positions, to ``map'' the aerodynamic forces added by the air duct.
  \textbf{B. Forces added by the air duct in a 40cm diameter circular air duct.} The arrow shows the direction, the ellipse the variance, and the color the magnitude of the force (N.). The displayed force is the measured force to which the forces outside the duct are subtracted.
  \textbf{C. Interpretation of the forces in a 40cm diameter air duct.} The blue zone corresponds to the ground effect. In the purple zone, the drone is pushed downward. In the red zones, it is ``sucked up'' by the walls, that is, these are unstable positions that are likely to lead to collisions. The green zone is the most stable one, as most effects are canceled. Counter-intuitively, the center of the duct is not the most stable position; instead, the airflow is less perturbed at an altitude of about 10 cm (above the ground effect).
  \textbf{D. Forces added by the air duct in a 50cm diameter air duct.} The pattern is similar to the one in the 40 cm diameter air duct.
  \textbf{E. Forces added by the air duct in a 50cm $\times$ 50cm square duct.} The arrow shows the direction, the ellipse the variance, and the color the magnitude of the force (N.). The displayed force is the measured force to which the forces outside the duct are subtracted.
  \textbf{F. Interpretation of the forces added by the air duct in a 50cm $\times$ 50cm square duct.} There is no downward force, but there are clear (1) ground effect (blue), wall effect (red), and ceiling effect (yellow). The center of the duct is stable, but the ceiling and the walls are unstable, as they pull the drone towards the borders. 
  }
\end{figure*}

Quadrotors are now ubiquitous in outdoor environments and fulfill many missions, from entertainment to movie-making and industrial inspections, but they are also increasingly being tested in underground environments. Many promising experiments are in mines \cite{li2020autonomous,elmokadem2022method}, which are hazardous and often require going up vertically. Beyond mines, the DARPA recently (2021) challenged research teams to propose technologies that can navigate human-made tunnel systems, urban underground, and natural cave networks, intending to provide situational awareness to a small team of operators  \cite{orekhov2022darpa,chung2023into}. In the final round, 11 out of the 17 teams deployed drones, in addition to other robots, including the winning team \cite{cerberus2022}. Overall, drones are appealing candidates for tunnel environments for the same reasons as for the smaller ducts and pipes: they can move vertically, from climbing ladders to negotiating vertical wells, and fly over complex ground terrains without being slowed down.

A few flying prototypes have been built to fly in manholes, penstocks and HVAC (Heating, Ventilation and Air-Conditioning) ducts, by mainly focusing on the tolerance to collision with variants of protective cages \cite{briod2014collision}. 
Hence, de Petri et al. \cite{de2021resilient} demonstrated a sub-500g drone that flies into a rectangular 0.5$\times$0.4 m-wide and 5.2 m-long manhole that involves a yaw turn, with flaps continuously touching the wall to prevent the shield from colliding with the manhole. In a larger diameter, a drone enclosed in a 40cm ``shell'' recently flew in a 1.5 m diameter penstock with a sloped section \cite{abayan2022tof}. Another drone, enclosed in a 40cm circular cage that rolls on the ground, moved into a rectangular 60$\times$75-cm-wide and 175-cm-long wooden HVAC duct prototype with its top side opened \cite{khalil2023autonomous}.


Several challenges arise when a drone attempts to fly in a highly confined space, like an air duct, that is typically not encountered in larger tunnels. First, the aerodynamic interactions between the rotors and the environments are complex and unknown because the airflow ``comes back'' to the drone and perturbs it significantly~\cite{wang2021estimation}. Second, these turbulences, combined with the proximity of the walls, make the position estimation highly critical; but this estimation is difficult in dark and almost featureless environments \cite{ozaslan2017autonomous,wang2022neither}. These challenges are amplified by the fact that the drone needs to be as small as possible to fit in the smallest ducts possible. 

To our knowledge, the only attempt to explicitly address these challenges was only recently published \cite{wang2022neither}, in which a 1.23 kg/40 cm drone flew autonomously in a 0.6 m air duct. The authors used computational fluid dynamics simulation to select the best flying speed, following the intuition that moving allows the drone to avoid flying in the turbulences it creates, but flying too fast makes maneuvering, wall avoidance, and control difficult.

Here, our objective is to address the specific challenge of \emph{hovering} a small quadrotor in a 35 to 50-cm air duct, considering that hovering is harder than translating~\cite{wang2022neither}, but mandatory for many applications like inspection. We use a 180 mm (including rotors) / 130 g quadrotor based on the Bitcraze Bolt, a derivative of the Crazyflie \cite{giernackiCrazyflieQuadrotorPlatform2017}, with brushless motors and additional distance sensors (Fig.~\ref{fig:concept}B, Methods). 

We first designed a robotic system to measure the forces that are created by the airflow when flying in air ducts. We obtained a ``map'' of the forces in a circular air duct. From these experiments, we found the position that is the less affected by air recirculations. We then designed a positioning system that makes it possible for the drone to hold its position, which is an input to a classic position loop. We used miniature time-of-flight (ToF) sensors combined with a neural network that estimates the position given the measurements. Putting these two components together, the optimized position and a reliable position estimate, we show that a 180 mm drone can hover autonomously in a 350mm air duct, which opens a whole new flight domain for drones.

\section*{Results}
\subsection*{Aerodynamic forces in air ducts}

When the air is accelerated by the four rotors, it is expected to hit the floor, then circulate and potentially go back to perturbate the drone from the top or the side. The most well-known of the expected effects is the ground effect \cite{cheeseman1955effect,shi2019neural}, which creates additional lift when a drone or a plane is close to the ground; but there are also ``ceiling effects'' \cite{conyers2018empirical} and ``wall effects'' \cite{robinson2014computational, tanabe2018multiple, hughes2021wall} and  ``corner effects'' \cite{prothin2019aerodynamics}, which are less explored. Interestingly, these surface effects can be used to detect obstacles in the vicinity of the drone \cite{ding2023aerodynamic, nakata2020aerodynamic}. To our knowledge, there is currently no data about the effects of rotors in circular pipes or air ducts, in which all these effects combine together.

We first observed that these effects substantially impact the capability of a drone in an air duct. In this experiment, a drone was controlled to hover inside and outside a 35-cm air duct, at the center. In both cases, its controller relies on an external, absolute position system (HTC Vive, see Methods). In free air, the drone is very stable, with typical oscillations of less than 5 mm around its target position, and not more than 20mm (Fig.~\ref{fig:concept}D, orange). By contrast, inside the duct, the drone has often an error of almost 60mm, and is generally much less capable of staying at the center (Fig.~\ref{fig:concept}D, green).

To understand the aerodynamic forces in a pipe in a systematic way, we designed a robotic setup based on a 7-degree of freedom manipulator (Franka-Emika Panda) and a force/torque sensor (Fig.~\ref{fig:aero}A) (Methods). Our objective is to measure the forces added by the circular duct on the drone for many positions. To do so, we fixed the quadrotor to the force/torque sensor and mounted it on a 3D-printed horizontal pole, which acts as the end-effector of the manipulator. For each measure, the quadrotor is started so that the lift compensates its weight (about 50\% of the maximal power with our quadrotor). This sets the reference for the force measurements in ``free air'', that is without any interaction with the environment. We then ask the robot to move the drone inside the duct and keep it inside for 10 seconds, measuring the forces and filtering them with a low-pass filter (Methods, Fig.~S\ref{si-fig:time}). We repeated this procedure for 192 regularly spaced points to obtain a ``map'' of the forces. For each point, we computed the mean force inside the duct, subtract the mean force outside of the duct, and computed the standard deviation along both axes to compute an ellipse that represents the uncertainty of the measurement (Methods).

The results show a coherent structure of the forces along the whole duct (Fig.~\ref{fig:aero}B). For a diameter of 40 cm, we observe a ground effect for the first 8 to 10 cm (Fig.~\ref{fig:aero}C, blue). Please note that we do not have data for positions closer to the ground because of the configuration of the pole that supports the drone and the possible risks of colliding with the duct, but we expect the ground effect to be more important when the drone gets closer to the ground. If we look at a vertical line centered in the duct, we observe an overall downward force, especially at the center of the duct  (Fig.~\ref{fig:aero}C, purple), which is canceled by the ground effect once close enough to the floor (Fig.~\ref{fig:aero}C, green). On a horizontal axis, the two top quadrants show forces mostly horizontal that attract the drone to the walls  (Fig.~\ref{fig:aero}C, red).

We observe the same pattern of forces for all the diameters of the circular ducts that we investigated, from 40 to 65 cm  (Fig.~\ref{fig:aero}D, Fig.~S\ref{si-fig:all-diameters}). In all circular pipes, the area with the least perturbation is around the bottom quarter of the pipe, horizontally centered. As expected, when the power is reduced, the forces keep the same direction but reduce in magnitude, and, conversely, they increase when the power increases (Fig.~S\ref{si-fig:all-motors}).

In a square air duct, the forces follow a different, simpler pattern (Fig.~\ref{fig:aero}E). There is a clear, vertical ground effect (Fig.~\ref{fig:aero}F, blue) up to about 15 cm (for a 50 $\times$ 50 cm air duct). Wall effects that attract the drone towards the wall are mostly horizontal and start at about 15cm from the wall. Interestingly, the ceiling effect pulls the drone up when it is closer than 15 cm from the ceiling, whereas this effect is not observed in circular ducts.

Overall, these maps of forces created by a drone flying in a pipe or air duct confirm the complexity of the airflow in this environment and explain why a standard drone has a much harder time flying in an air duct than outside. The exact value of the forces depends on many parameters, most noticeably the velocity, the lengths and the number of blades of the rotors, which, in turn, are linked to the size and mass of the drone, as well as many design choices. At the current stage, it is therefore challenging to define a general model that would give an estimate of the forces for any kind of drone. 

Nonetheless, the consistent pattern of forces across varying duct diameters and motor regimes enables the formulation of qualitative guidelines for the best drone stability. In a circular air duct, contrary to intuition, optimal stability is achieved by flying at a height of about 15 cm (about a quarter of the duct diameter in a 40 cm duct). This positioning places the drone above ground effects yet below downward forces, effectively stabilizing its position (green zone of Fig.~\ref{fig:aero}). The top left and top right quadrants pose instability risks due to the drone's attraction towards the duct borders, rendering precise position control challenging. By contrast, in a square air duct, the center is the less impacted zone and attraction forces become significant if the drone gets closer than about 15~cm of the wall or ceiling. Notably, while a circular duct's ceiling does not attract the drone, a rectangular duct's ceiling poses a safety hazard because it pulls the drone up.

\subsection*{Data-driven Localization}
From our aerodynamic experiments, we deduce an optimal placement of the drone in an air duct. However, even at the best position, the drone must contend with the turbulences while maneuvering in close proximity to the walls that surround it in every direction. In such a situation, precise position control is key, as even minor deviations of a few centimeters can lead to a collision which will likely be followed by a crash. The primary challenge for position control lies in accurately estimating the drone's position within the duct: once the position is known, conventional position control methods such as PID \cite{aastrom2006advanced} or more sophisticated techniques \cite{song2023reaching} can be implemented.

A drone in an air duct can hardly exploit vision-based localization techniques, like visual SLAM \cite{macario2022comprehensive} or optic-flow \cite{serres2017optic}. First, its environment is typically featureless, almost uniform, whereas these techniques rely on salient visual features and contrasts. Second, as air ducts are dark, the drone must provide its own illumination, whereas onboard lights cannot be powerful on a small drone without impacting the autonomy significantly and do not cast light uniformly. Last, a sub-150g drone cannot embed much computing power, making it challenging to work with image processing algorithms.


Instead, our drone exploits 9 miniature time-of-flight sensors (Methods), one toward the ground, and 8 in the horizontal plane; these sensors have a range of 4 meters and do not require any light. 

When we assume that the drone is horizontal and the time-of-flight sensors are casting a single ray, the altitude (z position) corresponds to the distance returned by the sensor that is directed toward the ground, and the horizontal position in a slice of the duct (y position) can be solved analytically with a geometrical method (Methods). 

Nonetheless, the current time-of-flight sensors we use have a typical field of view of 27 degrees (Methods). As a consequence, when a sensor is not perpendicular to the wall, that is, when the rays are ``tangent'' to the walls, it will return some average distance that is not fully determined. Please note that this is always the case for a subset of the sensor for any position of the drone.

\begin{figure*}
  \includegraphics*[width=\textwidth]{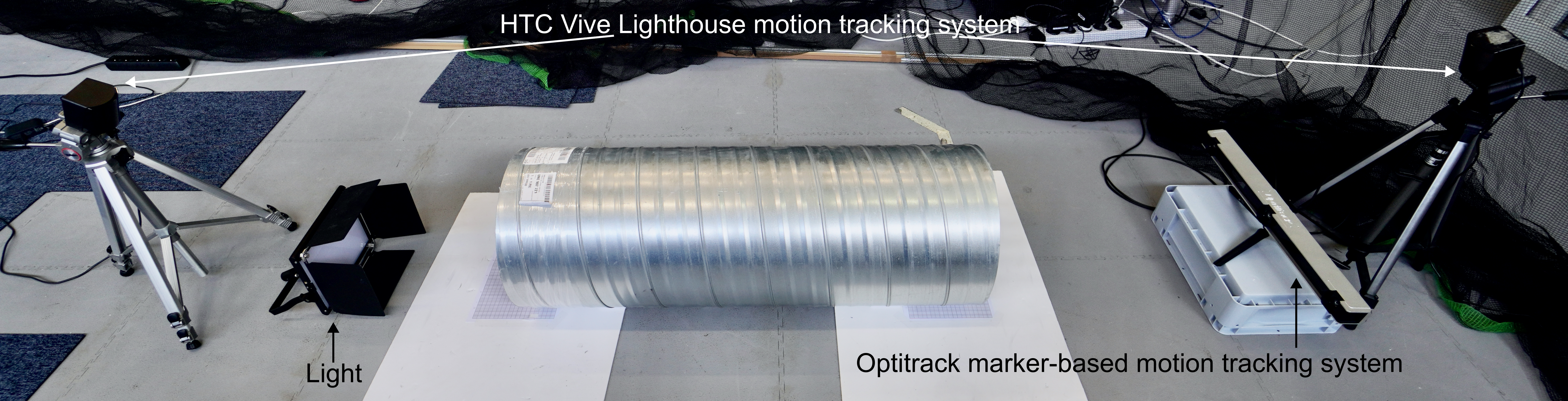}
  \caption{\label{fig:setup-dataset}\textbf{Experimental setup to acquire data for the data-driven localization.} From left to right, a Lighthouse base station (on a tripod), an LED panel light, a circular air duct in which the drone flies, an Optitrack Trio, and a Lighthouse base station (on a tripod).}
\end{figure*}

While the sensor properties could be measured and modeled depending on their angle with the wall, we decided to follow a more ``end-to-end'' approach by learning a neural network that computes the horizontal and vertical positions (y and z) given the horizontal and vertical velocities, and roll and pitch angles, and the laser measurements. Velocities were introduced after testing various parameter combinations, leading to improved experimental flight performance. The yaw input was omitted, assuming the drone's forward axis aligns with the air duct's axis. This decision also avoids potential issues that could arise in the neural network if the yaw angle falls outside the [-90°, 90°] range, a concern not applicable to the roll and pitch angles. Our objective is for this model to take into account both the inaccuracies of the sensors but also implicitly disambiguate computations by learning priors based on the statistical properties of typical flights.

This neural network is learned with classic supervised learning. To do so, we need to acquire the ground truth, here the full pose of the drone, with a reliable external system; this is challenging because the duct walls hide the drone from standard optical tracking systems (e.g. Optitrack or Vicon). We considered two options: (1) Optitrack Duo/Trio, which is a system based on two/three calibrated infrared cameras that can be placed in the axis of a duct, and (2) the HTC Vive lighthouse system \cite{niehorster2017accuracy}, which relies on a rotating IR laser in one or two external beacons. We decided to exploit the HTC Vive system because the position is computed internally by the Crazyflie drone \cite{taffanel2021lighthouse}, which makes it easy to align temporally the measurements from the lasers, the IMU, and the ground truth (as they are all computed in the same system). By contrast, external systems like Optitrack need to send the position by radio to the drone which adds at least a few milliseconds of delay. The setup is made of a one-meter segment of air duct placed horizontally on the ground with 2 Lighthouse base station v2 situated approximately 60 cm from each extremity of the duct (Fig.~\ref{fig:setup-dataset}). In the duct, the drone is controlled remotely by a pilot using a gamepad to have more control when flying close to the duct. Using this approach, we acquired a dataset of 9 flights, totaling 33.9 minutes, at a frequency of 250 Hz. A data augmentation process utilizes the XZ plan of symmetry to double the training dataset size. Overall, the dataset is made of 891614 points, divided into a training set of 784836 points and a test set of 106778 points.

The position estimator is a multilayer perceptron (MLP) with 4 layers, for a total of 63 neurons (Methods). It takes 13 inputs (the 9 time-of-flight sensors, the y and z velocities and the roll and pitch angles estimated by the Kalman filter from the IMU). The inputs pass through 2 hidden layers of size 32 and 16. The outputs are the y and z positions. It is trained on a computer with the PyTorch library and then integrated into the firmware of the drone using the X-CUBE-AI toolkit from STMicroelectronics. It runs at 10 Hz on the onboard STM32F4 microcontroller (Methods).

\begin{figure*}
  \includegraphics*[width=\textwidth]{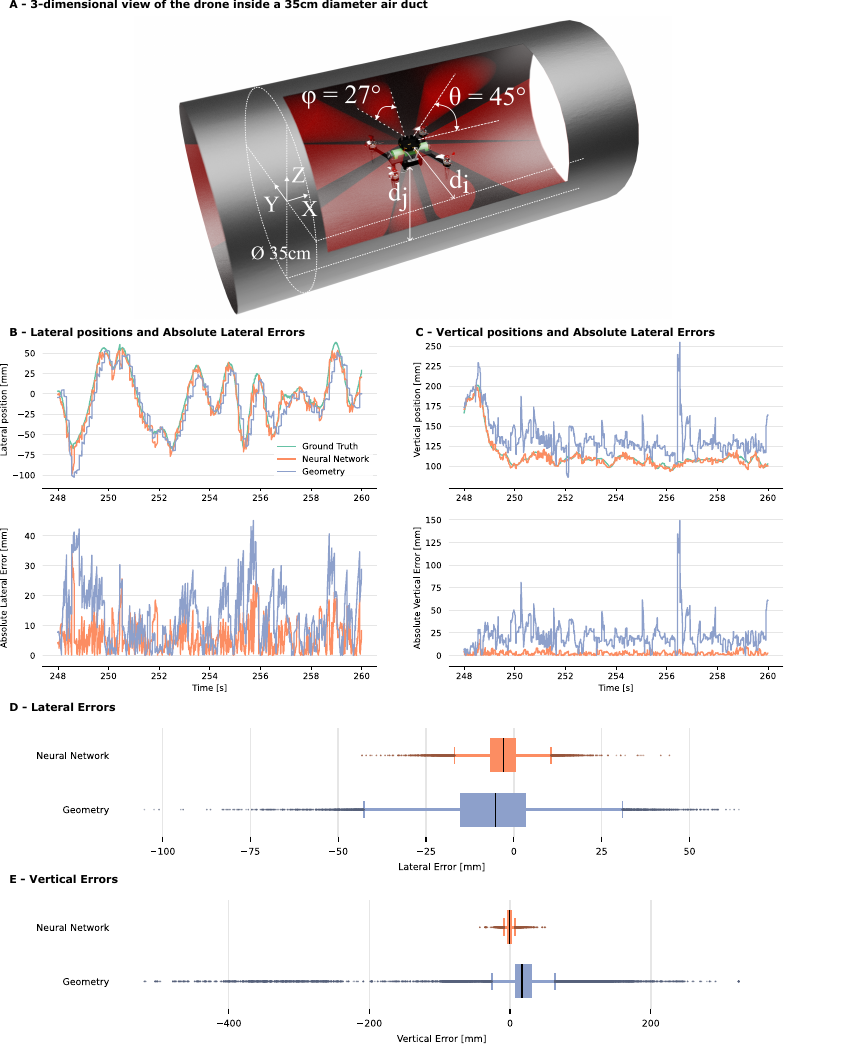}
  \caption{\label{fig:simul_results}
  \textbf{A. 3D view of the drone flying inside a 35cm diameter air duct.} The drone is in the middle of the cylinder. The emission cone of the time-of-flight (ToF) sensors is represented in red and is $\phi = 27^\circ$. ToF sensors emitting on the horizontal plane are separated by $\theta = 45^\circ$. $d_i$ and $d_j$ represent 2 distances measured by the ToFs.
  \textbf{B-E. Comparison of the position estimations from the geometric and neural network methods against the ground truth inside a 35 cm air duct.} The green line represents the Ground Truth measured by the Extended Kalman Filter of the drone helped with the Lighthouse positioning system. The orange and blue lines represent respectively the lateral position outputted by the neural network and the geometrical solution.
  \textbf{B. Lateral positions and Absolute Lateral Errors.} In this time section of the test set, the geometrical solution is less accurate than the neural network.
  \textbf{C. Vertical positions and Absolute Vertical Errors.} In this time section of the test set, the geometrical solution is less accurate than the neural network.
  \textbf{D. Lateral Errors.} This boxplot represents the dispersion of the lateral position error outputted by the neural network and the geometrical solutions with respect to the ground truth in the test set. The neural network outputs a lateral position that is more precise and accurate than the geometrical solution.
  \textbf{E. Vertical Errors.} This boxplot represents the dispersion of the error between the vertical position measured by the neural network and the ground truth in the test set. The neural network outputs a vertical position that is more precise and accurate than the geometrical solution.}
\end{figure*}

The results on the test set (Fig.~\ref{fig:simul_results}) show that both methods (geometric and neural network) can estimate the position in the air duct. Nonetheless, the neural network is significantly more accurate than the analytical approach, with a median error of -3.0 mm versus -5.2 mm laterally (Fig.~\ref{fig:simul_results}D, [5\%,95\%] confidence interval [-13.3, 5.9] vs [-67.3, 30.1]), and -0.5 mm versus 17.1 mm vertically (Fig.~\ref{fig:simul_results}E, [5\%,95\%] confidence interval [-6.6, 5.1] vs [-67.3, 107.0]). The horizontal errors of the geometrical method are likely to originate from an imperfect measurement model for the sensors, although more experiments would be needed to understand this in detail. Importantly, the geometric computations sometimes have large errors that would directly result in a crash, in particular about 10 cm on the vertical axis for the 95\% confidence interval.

Additional data for 45 cm and 56 cm tubes are provided in Supplementary Material (Fig.~S\ref{si-fig:simul_results_45cm} and Fig.~S\ref{si-fig:simul_results_56cm}), which support the same conclusions.

\subsection*{Flight tests}
We uploaded the neural network to the onboard microcontroller (Methods, Fig.~\ref{fig:schema_estim}) and used its outputs as the inputs of the Kalman filter used by the Crazyflie software to estimate its position, velocity and orientation. In concrete terms, we plug the neural network as if it were a motion capture system (like the HTC Vive system that we use for training). As a consequence, the output of the neural network is merged with the IMU, the optical flow sensor and the model of the drone, and not used directly. The neural network does not estimate the ``forward'' (x-axis) position, which is, in our opinion, an open problem of navigation that is unrelated to aerodynamics \cite{elmokadem2022method,delgado2015vision,garrote2024exploiting}. To perform the experiments, an LED panel light was positioned on the ground in front of the air duct entrance. The light is oriented horizontally, slightly towards the ground to produce a grazing light shining on the spiral bump of the duct to cast shadows, thereby enhancing the visual cues available to the drone's optical sensor to estimate the ``forward'' position.


We used the same air duct as the one used for training, that is, we consider that a neural network is trained for each duct diameter and that the operator of the drone can easily select the right network. We did not consider changes in diameter during flight since we were focused on hovering and not navigation. However, our preliminary tests show the system is robust to different diameters as depicted in Fig.~S\ref{si-fig:supplementary_stats}. To get an independent ground truth, we set up an external marker-based visual tracking system (Optitrack V120:Trio, Methods) that is both independent from the onboard estimation and the system used for the dataset used in the training phase. 

\begin{figure*}
  \includegraphics*[width=\textwidth]{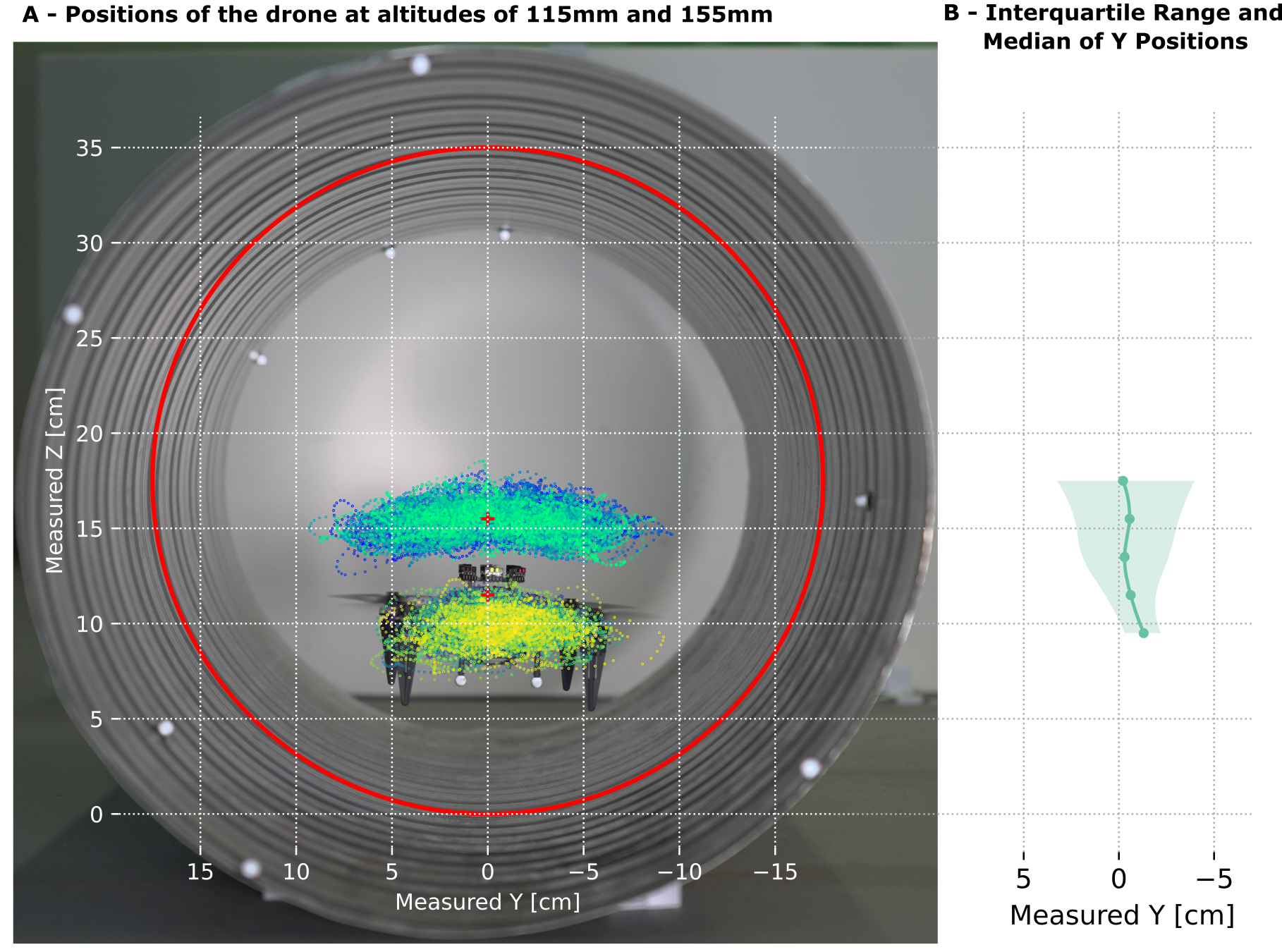}
  \caption{\label{fig:flight_results}
  \textbf{A. Positions of the drone at altitudes of 115mm and 155mm.} Two 2-minute flights are shown in this plot. The red circle represents the 35-cm air duct. The 2 red crosses depict the position the drone must keep for each flight where the drone takes off and hovers at an altitude (Z) of 115mm or 155mm in this air duct. It must stay at $Y = 0$. Two groups of points are visible. The group located below (colors from blue to yellow) are the positions of the drone for the flight at 115mm. The group located above (colors from blue to green) represents the positions of the drone for the flight at 155mm. The neural network used is the one trained on this 35-cm air duct.
  \textbf{B. Interquartile Range and Median of Y Positions.} This plot depicts the interquartile range (in light green color) and the median (the bright green line) of the lateral positions taken by the drone for different flight altitude targets (represented by the green dots). The target altitude is on the ordinate and the abscissa represents the measured lateral position. It uses the same neural network as the figure A on the left.}
\end{figure*}

We first performed a 2-minute flight at an altitude of 11.5 cm in a 35-cm circular air duct (Fig.~\ref{fig:flight_results}A). Overall, the drone is stable enough to fly, which shows that flying an 18-cm drone in a 35-cm air duct is possible by combining both the choice of altitude and the data-driven position estimator.

We then investigated if, as predicted by our aerodynamic forces measurements (Fig.~\ref{fig:aero}), the drone is less stable at the center of the duct than at the recommended altitude. To do so, we performed stationnary flights at 5 altitudes separated by 2 cm between 9.5 cm (the lowest experimentally possible) and 17.5 cm (the center of the duct and the highest experimentally possible) and observed the deviations from the target position. These results (Fig.~\ref{fig:flight_results}B) confirm our aerodynamic force measurements: the drone is significantly less stable when the altitude increases, with an amplitude of oscillation (first quartile) of about 1 cm at an altitude of 9.5 cm versus 3.6 cm at~17.5 cm.

Additional data for 45 cm and 56 cm tubes are available in Supplementary Material. They show similar results. Incidentally, the neural network generalizes well enough to enable the quadrotor to fly in a duct with a diameter 10 cm smaller or larger than the one on which it has been trained (Fig.~S\ref{si-fig:supplementary_stats}).

\section*{Discussion}
\subsection*{A new domain for flying robots}
Our findings demonstrate that a quadrotor weighing less than 150 grams can effectively hover within a 35-centimeter diameter air duct by combining (1) an altitude below the center of the duct, as showed by our aerodynamical forces study, and (2) a neural-network position estimator trained through an ``end-to-end'' approach. This success opens a whole new domain for flying robots --- air ducts and similar pipes -- with potential applications that span from industrial inspection to public safety.

One of the major challenges for future applications will be dust, which is a common occurrence in most air ducts, particularly on their lower surfaces. To mitigate this issue, the drone needs to move less air or at a lower speed, which is typically achieved by designing it to be lighter and by increasing the size of the rotors. More experiments are needed to evaluate the extent of the issue, as typical air duct networks are engineered to prevent dust buildup, in particular, by incorporating air filters \cite{batterman1995hvac}.

A second challenge for applications will be the navigation in air duct networks, encompassing progression through the network, selecting appropriate junctions, and subsequently returning to the entrance with the recorded images. As mentioned before, SLAM algorithms are successful in robotics \cite{siciliano2016robotics,macario2022comprehensive,ebadi2023present} but they are unlikely to work well in a featureless environment like an air duct, and they are computation-intensive. Visual odometry based on optic flow would be easier to embed \cite{floreano2013miniature, bergantin2023indoor}, but is also challenging in a dark environment that is only lit with small embedded lights \cite{castillo2023synchronization}. Teleoperation, that is, using a pilot, would be a possible approach but maintaining a radio link is challenging in underground tunnels \cite{rak2007uhf, boutin2008radio}; it may still be a possible solution in metal ducts \cite{nikitin2003propagation, bangar2021evaluating}. However, more research is needed in radio transmission inside complex metal duct systems over a long range with bends, junctions and the presence of obstacles and materials that could dampen the signal. A solution could be using drones to physically carry data (data muling) when important data is needed from the ground station or the main drone \cite{tuyishimire2017cooperative}. A promising research avenue is to maintain this radio link using a chain of drones that propagate the signal from drone to drone \cite{laclauSignalBasedSelfOrganizationChain2021}.

\subsection*{Control}

For this study, we used a conventional position controller based on a cascade of PID controllers \cite{giernackiCrazyflieQuadrotorPlatform2017}. We performed a preliminary study in simulation using model-predictive control instead of this position controller, so that the quadrotor could anticipate the forces and include its control decisions \cite{vladthesis}. However, our results did not show an improvement significant enough to justify the complexity of implementing real-time model-predictive control on such a light quadrotor. Simpler schemes based on feed-forward terms \cite{guzman2024tuning} in the PIDs could be enough to take the knowledge of the forces into account. Nevertheless, our measurements indicate that (1) the forces fluctuate significantly over time, which means that they are unlikely to be perfectly predicted, and (2) they are small enough to be ignored when the drone flies at the recommended altitude and position. Our current conjecture is that explicitly integrating force predictions into the position controller may only prove advantageous when navigating in close proximity to the ceiling or the upper sides of the duct.

Instead of improving the control law, non-coaxial drones \cite{rashad2020fully} could give more command authority and allow the drone to compensate directly in the horizontal plane, without adding roll or pitch motions. To our knowledge, no small non-coaxial drone has been designed yet, but this is a promising research avenue.

\subsection*{Aerodynamics}

In order to comprehend the aerodynamic phenomena at play, we designed a robotic setup that measures the average forces at different positions in a duct. While these experiments offer valuable insights and propose an airflow pattern, the system overlooks the drone's motion, as it remains fixed. In particular, a controlled quadrotor continuously adjusts its attitude to compensate for turbulences, but, by doing so, it changes the direction of the flow accelerated by the rotors, which, in turn, changes how it circulates. This closed-loop system, which can, for instance, lead to the amplification of control instabilities, would require more experiments. 

Furthermore, our setup only indirectly measures the airflow within the duct by assessing how the duct modifies the forces exerted on the quadrotor. 
Modern techniques like Particle Inverse Velocity \cite{adrian1991particle,cavazzini2012particle} offers the potential to directly visualize and possibly model the air velocity field, enhancing our understanding of its dynamics.
Specifically, it would be useful to know if there are temporal patterns of turbulence, or chaotic elements that cannot be modelled. We expect these future experiments in aerodynamics to lead to numerous insights about the best strategy to stabilize a drone in this specific, but well-defined, environment.


\section*{Author contributions}
JBM led the study and analyzed the aerodynamic data. LR designed the quadrotor and the robotic setup for aerodynamic investigations, and performed the aerodynamic experiments. VT performed the aerodynamic experiments and the ``geometry-based'' localization method. AG designed the data-driven localization method and trained the neural networks. TM performed the flight experiments and analyzed them. All authors analyzed the results and wrote the article.

\section*{Acknowledgments}
This work received funding from the DGA-Inria project "U-Drone", a CDSN PhD scholarship, the ANR/DGA ASTRID project Proxilearn (ANR-19-ASTR-0009), the ANR/DGA ASTRID project SpotReturn (ANR-21-ASRO-0001), and the ANR project BUCOLYC (ANR-23-CE51-0037).

\section*{Data availability}
The data from the aerodynamic experiments is available on Zenodo (\url{https://dx.doi.org/10.5281/zenodo.13837470}). The data to train the neural networks is also available on Zenodo (\url{https://dx.doi.org/10.5281/zenodo.13872875}), with the flight experiments data.

\section*{Code availability}
The Python scripts to reproduce the analyses are included with the data. The C code of our quadrotor, based on the bitcraze open-source code, will be available on \url{https://github.com/hucebot/}.

\section*{Methods}
\subsection*{Drone}
The drone (Fig.~\ref{fig:concept}B) is 180 mm wide (including rotors) and 75 mm high (feet and an additional deck on top included). It weighs 130 g (battery included). The main board is a Bitcraze Bolt, a derivative of the Crazyflie \cite{giernackiCrazyflieQuadrotorPlatform2017}, based on an STM32F4 microcontroller at 168 Mhz. The main frame is custom-made in carbon fiber. The drone is powered by a single (type 18650) Li-Ion battery of 3.6V / 3 Ah, which gives an autonomy of about 15 minutes. The four brushless motors are the ROBO 1202.5 11500kv from Flywoo associated with GEMFAN 3018 2-blade propellers measuring 7.6 cm in diameter. The four ESCs are XSD 7A ESC from FlashHobby.

On the bottom of the drone, the Micro SD card deck by Bitcraze was added to collect the data for the training of the neural network. The ``Flow Deck v2'' by Bitcraze was added to add the downward-facing time-of-flight and an optic flow sensor.

On the top of the drone, a custom deck was added to add 9 time-of-flight sensors (STM32's VL53L1X), based on the design of the Multi-ranger deck by Bitcraze (which is similar, but with 5 sensors only).

\subsection*{Comparison of flight stability inside and outside a circular duct (Fig.~\ref{fig:concept})}
The drone flies for 2 minutes inside the 35-cm air duct and outside (Fig.~\ref{fig:concept}D). The drone's state estimator uses the HTC Vive's Lighthouse Positioning system as the position reference. The altitude command is 127mm for both flights so the propellers, located 50mm over the feets of the drone are about the middle of the duct. The data have been centered around the median. Y is the lateral position and Z is the altitude. X (not shown here) is the forward position along the tube axis.

\subsection*{Force measurements (Fig.~\ref{fig:aero}, Fig.~S\ref{si-fig:time}, Fig.~S\ref{si-fig:all-diameters}, and Fig.~S\ref{si-fig:all-motors})}
The drone is mounted on a 6D force/torque sensor (FTN-NANO17 with the calibration SI-12-0.12, by ATI technologies), which is connected to the acquisition computer by ethernet. According to the manufacturer, the sensitivity of the sensor are:
\begin{itemize}
\item Fx/Fy/Fz: 1/320 N
\item Tx/Ty/Tz: 1/64 N.mm
\end{itemize}
The data are acquired at 7 Khz.

The sensor is mounted on a 3D-printed pole, designed to be as rigid as possible, which is screwed in place of the robot end-effector (Fig.~\ref{fig:aero}).

For each measurement, the robot (Franka-Emika Panda) starts outside of the duct. The motors are started and the drone is kept outside for 5 seconds. Except for the data of Fig.~S\ref{si-fig:all-motors}, the motors are set to 50\% of their maximum value, which compensates for the weight of the drone. This gives the reference force/torque. Then, the robot puts the drone inside the air duct (as deep as possible given the workspace of the robot, which corresponds to roughly putting the center of the drone 30 cm inside).

The data are first filtered with a low-pass Butterworth filter of order 4 (from the \verb|scipy.signal| Python package), with a sampling frequency of 7 Khz with a cutoff frequency of 1 Hz.

On plots (e.g., Fig.~\ref{fig:aero}), the arrow corresponds to the mean of the force Fz and Fy, and the ellipse corresponds to the covariance ellipse (1 standard deviation).

\subsection*{Geometric position estimation (Fig.~\ref{fig:simul_results})}
Knowing the position of the drone whose center is $O_B$ in the cylinder, for each ToF $i$, the point $O_i$, which is the intersection between the ToF direction and the cylinder verifies the following equation: $O_{i,y}^2 + O_{i,z}^2 = r^2$. However the point $O_i$ is the translation of the point $O_B$ by the orientation vector of the ToF: $\mathbf{d_i} = d_{i}\mathbf{s_i}$, hence its coordinates based on the position of the drone

\begin{equation}
\left\{ 
    \begin{array}{l}
      O_{i,x} = O_{B,x} + d_{i}s_{i,x}\\
      O_{i,y} = O_{B,y} + d_{i}s_{i,y}\\
      O_{i,z} = O_{B,z} + d_{i}s_{i,z}
    \end{array}
\right.
\end{equation}







$W$ and $B$ are respectively the world and drone coordinate frames and $R_{WB}$ is the rotation matrix to pass from the drone to the world coordinates. This matrix is known from the gyroscope of the drone.
\begin{equation}
[\mathbf{s_i}]_{W} = R_{WB}[\mathbf{s_i}]_{B}
\end{equation}

We have for 2 distinct ToF $i$ and $j$:
\begin{equation}
\begin{split}
  r^2 &= (O_{B,y} + d_{i}s_{i,y})^2 + (O_{B,z} + d_{i}s_{i,z})^2\\ &= (O_{B,y} + d_{j}s_{j,y})^2 + (O_{B,z} + d_{j}s_{j,z})^2 
\end{split}
\end{equation}

Which gives us
\begin{multline}
2O_{B,y}(d_is_{i,y}-d_js_{j,y}) + 2O_{B,z}(d_is_{i,z}-d_js_{j,z})\\ = d^{2}_j(s_{j,y}^2 + s_{j,z}^2) - d^{2}_i(s_{i,y}^2 + s_{i,z}^2)
\end{multline}

with the following notation:
\begin{equation}
\begin{array}{l}
  a_{k} = a_{i,j} = d^{2}_j(s_{j,y}^2 + s_{j,z}^2) - d^{2}_i(s_{i,y}^2 + s_{i,z}^2)\\
  b_{k} = b_{i,j} = 2(d_is_{i,y} - d_js_{j,y})\\
  c_{k} = c_{i,j} = 2(d_is_{i,z} - d_js_{j,y})
\end{array}
\end{equation}

We get the equation
\begin{equation}
\begin{array}{l}
a_k = b_kO_{B,y} + c_kO_{B,z}\\
A = (B|C)\left(\begin{array}{l} O_{B,y}\\ O_{B,z}\end{array}\right)
\end{array}
\end{equation}

A, B and C are column vectors of the $a_k$, $b_k$, and $c_k$ coefficients. If the number of ToF is $k>2$ the position is the solution of the root mean square problem. The solution can be calculated with the pseudo-inverse of the matrix $M=(B|C)$
\begin{equation}
\left(\begin{array}{l} O_{B,y}\\ O_{B,z}\end{array}\right) = (M^{T}M)^{-1}M^{T}A
\end{equation}

\subsection*{Training of the neural network (Fig.~\ref{fig:simul_results})}
During data-gathering flights, we utilized the Bitcraze SD-deck to record data at 250 Hz, including ToF ranges and state estimator outputs such as position, speed and attitude. The drone is stabilized using a cascaded PID controller and accurate localization is performed using two HTC Vive Lighthouse Base Stations as part of the global positioning system. This configuration helps the drone to fly close to walls by a pilot using a Logitech F710 gamepad. However, the Lighthouse deck extension used to get the localization from the HTC Vive system masks the ToF sensor measuring the distance on top of the drone. Hence, the data coming from this sensor are discarded. To prepare the data for neural network training, we first remove all segments containing crashes. Additionally, since the setup comprises a one-meter segment of an air duct and to prevent laser sensors from measuring obstacles outside the duct, any laser range above 50 cm is replaced by zero. This dataset consists of 9 flights, averaging 3.15 minutes per flight and totaling 29.4 minutes. Finally, a data augmentation process utilizes the XZ plan of symmetry to double the dataset size.

\subsection*{Implementation of the neural network in the firmware}
\begin{figure*}
  \centering
  \includegraphics*[width=\linewidth]{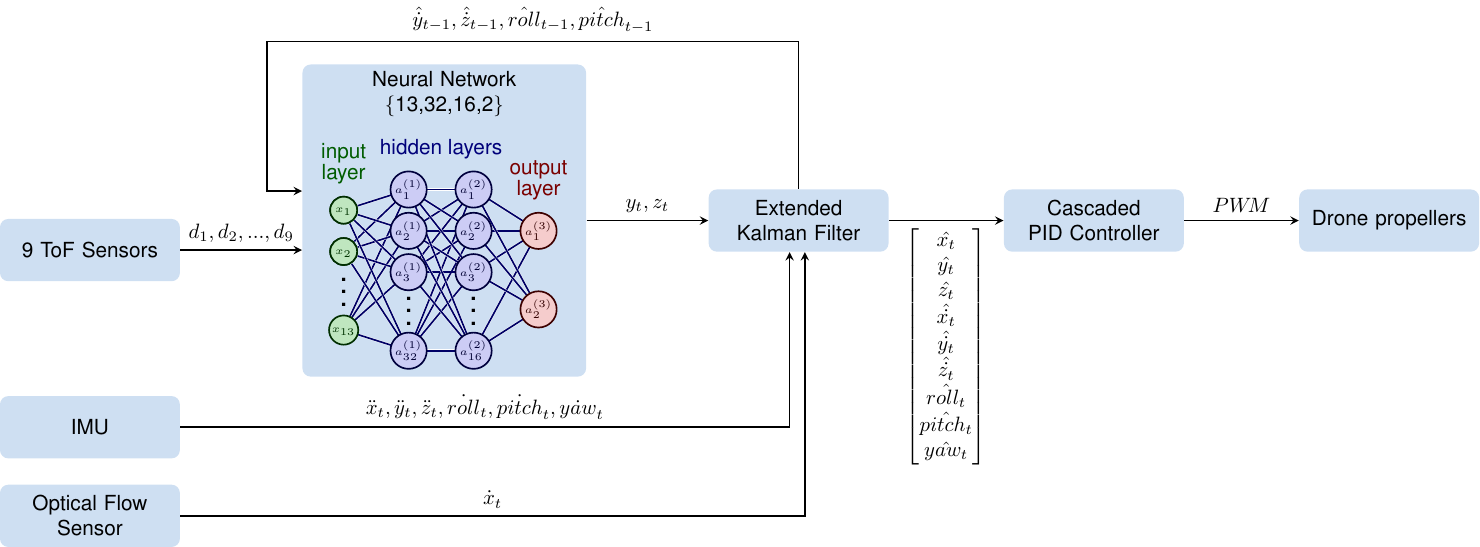}
  \caption{\label{fig:schema_estim}\textbf{The system architecture.} The neural network uses as input the velocities $\hat{\dot{y}}_{t-1}$ and $\hat{\dot{z}}_{t-1}$ and the angles $\hat{roll}_{t-1}$ and $\hat{pitch}_{t-1}$ previously estimated by the EKF (Extended Kalman Filter) as well as 9 ToF Sensors distances. The output is an estimation of the $y$ and $z$ position of the drone in the circular section of the duct. The EKF takes as inputs the outputs of the Neural Network, the IMU and the Optical Flow Sensor to output an estimate of the position, velocity and angle of the drone. A Cascaded PID controller is then used to control the drone propellers, comparing the estimated dynamics of the drone against the programmed commands.}
\end{figure*}

We use the X-CUBE-AI toolkit v6.0.0 from STMicroelectronics to convert the weights of the neural network into optimized C code that can run on the MCU of the drone. The choice of this toolkit is motivated mainly by its ease of use. Also, its low memory consumption and low latency performances \cite{osman2021tinyml} are key factors to enable running a neural network alongside the existing algorithms on the system. The output of the neural network is then inputted into the Extended Kalman Filter (EKF) of the drone as depicted in Fig.~\ref{fig:schema_estim}. This neural network gets inputs at a rate of 10 Hz. The inference time is about 0.2 ms on average. The controller is the state-of-the-art PID cascaded controller of the Bitcraze Bolt.

\subsection*{Experimental setup for flight tests (Fig.~\ref{fig:setup-dataset})}
The V120:Trio camera from Optitrack is placed at about 60 cm of one extremity of the duct so at least 2 cameras can track the drone and the duct. The selection of this camera is based on its millimeter-level precision in tracking small objects like drones, as demonstrated in \cite{martin2024compact}. The camera is close enough to track 6.4-mm-diameter reflective markers stuck inside the cylindrical duct and on a 3d printed structure fixed on the rear of the drone. The extensions on the drone are the SD-card deck to plug a micro SD-card to collect data from the drone, the Flow deck v2 containing a ToF (VL53L1X) and an optic flow sensor, and the MultirangerX9 deck containing 9 ToF (VL53L1X) (the one measuring the top distance is not used as explained in Methods). The tube is lit with an LED spot close to the ground so the spiral structure of the duct can cast some shadow and the reflection of the light on the tube does not saturate the optical sensor. This setup enables the possibility of using the Flow deck in the X direction (forward) to estimate the last coordinate position not calculated by the neural network.

The drone is placed in the middle of the duct with its forward direction aligned with the axis of the duct before take off. The drone takes off and hovers for 2 min at a given altitude while keeping its lateral position in the center of the tube. Then, the drone lands. The altitude is increased incrementally until the drone is too unstable to fly in the duct.

\subsection*{Calibration of data for analysis}
The positions of the markers stuck inside the duct and measured by the Optitrack are used to fit a cylinder to align the X vector of the reference frame with the axis of the duct. Then, the reference frame of the Optitrack data are translated to the middle of the bottom of the tube. This attitude and position change of the reference frame is reflected in all Optitrack data post-flight.

The neural network implemented on the drone has been trained with the Lighthouse deck which sets an altitude of 0 cm when the drone is positioned on the ground. The position of the drone measured by the Optitrack is located in the center of the drone 3.5 cm above the ground when it is not flying, where the control board is. Consequently, the target altitudes are shifted 3.5 cm up in the Y-axis of the final plots.

The 10 first seconds after take-off detection and the last 5 seconds before landing are removed in the final dataset which gives us 5 $\times$ 1min 45s of steady state flight equaling 8min and 45s of flight data for each altitude shown in Fig.~\ref{fig:flight_results}B-C and Fig.~S\ref{si-fig:supplementary_stats}.

\subsection*{Flight results for other air duct diameters (Fig.~S\ref{si-fig:supplementary_stats})}
The same methodology explained precedently for the 35-cm tube has been used for the other air ducts (45 and 56 cm). Fig.~S\ref{si-fig:supplementary_stats} illustrates the flight results conducted in air ducts of 35, 45 and 56 cm using the neural network trained in these conducts. In these figures, the real altitude at which the drone must fly, or flight altitude target, is in most cases shifted by several centimeters depending on the neural network used. On the contrary, the median of the Y position errors is more vertical and at most 1.5 cm from 0.

In Fig.~S\ref{si-fig:supplementary_stats}A to D and G to J the drone flies with a neural network trained in a different air duct size.

In Fig.~S\ref{si-fig:supplementary_stats}E, G, I and K, the drone flight shows an increase in the interquartile range on the Y position as the altitude increases in the 45 and 56 cm air duct. This could explain an increase in aerodynamic turbulences on the drone. There is a slight increase in the size of the interquartile range of the Y error in the lower part of the 45-cm and 56-cm duct which could be the impact of the ground effect on the drone. However, this does not appear clearly when the drone flies into 35-cm duct (In Fig.~S\ref{si-fig:supplementary_stats}A and C).

\printbibliography

@inproceedings{giernackiCrazyflieQuadrotorPlatform2017,
  title = {Crazyflie 2.0 Quadrotor as a Platform for Research and Education in Robotics and Control Engineering},
  booktitle = {2017 22nd {{International Conference}} on {{Methods}} and {{Models}} in {{Automation}} and {{Robotics}} ({{MMAR}})},
  author = {Giernacki, Wojciech and Skwierczynski, Mateusz and Witwicki, Wojciech and Wronski, Pawel and Kozierski, Piotr},
  date = {2017-08},
  pages = {37--42},
  publisher = {{IEEE}},
  location = {{Miedzyzdroje, Poland}},
  doi = {10.1109/MMAR.2017.8046794},
  url = {http://ieeexplore.ieee.org/document/8046794/},
  urldate = {2022-04-12},
  isbn = {978-1-5386-2402-9}
}

@article{laclauSignalBasedSelfOrganizationChain2021,
  title = {Signal-{{Based Self-Organization}} of a {{Chain}} of {{UAVs}} for {{Subterranean Exploration}}},
  author = {Laclau, Pierre and Tempez, Vladislav and Ruffier, Franck and Natalizio, Enrico and Mouret, Jean-Baptiste},
  date = {2021-04-23},
  journaltitle = {Frontiers in Robotics and AI},
  shortjournal = {Front. Robot. AI},
  volume = {8},
  pages = {614206},
  issn = {2296-9144},
  doi = {10.3389/frobt.2021.614206},
  url = {https://www.frontiersin.org/articles/10.3389/frobt.2021.614206/full},
  urldate = {2022-04-11}
}

@inproceedings{tanise2017development,
  title={Development of an air duct cleaning robot for housing based on peristaltic crawling motion},
  author={Tanise, Yuki and Taniguchi, Kosuke and Yamazaki, Shota and Kamata, Masashi and Yamada, Yasuyuki and Nakamura, Taro},
  booktitle={2017 IEEE International Conference on Advanced Intelligent Mechatronics (AIM)},
  pages={1267--1272},
  year={2017},
  organization={IEEE}
}

@inproceedings{wang2006autonomous,
  title={Autonomous air duct cleaning robot system},
  author={Wang, Ya and Zhang, Jianhua},
  booktitle={2006 49th IEEE International Midwest Symposium on Circuits and Systems},
  volume={1},
  pages={510--513},
  year={2006},
  organization={IEEE}
}

@inproceedings{jeon2012design,
  title={Design of an intelligent duct cleaning robot with force compliant brush},
  author={Jeon, Seung Woo and Jeong, Wootae and Park, Duckshin and Kwon, Soon-Bark},
  booktitle={2012 12th International Conference on Control, Automation and Systems},
  pages={2033--2037},
  year={2012},
  organization={IEEE}
}

@article{roh2005differential,
  title={Differential-drive in-pipe robot for moving inside urban gas pipelines},
  author={Roh, Se-gon and Choi, Hyouk Ryeol},
  journal={IEEE transactions on robotics},
  volume={21},
  number={1},
  pages={1--17},
  year={2005},
  publisher={IEEE}
}

@inproceedings{shao2015review,
  title={A review over state of the art of in-pipe robot},
  author={Shao, Lei and Wang, Yi and Guo, Baozhu and Chen, Xiaoqi},
  booktitle={2015 IEEE International Conference on Mechatronics and Automation (ICMA)},
  pages={2180--2185},
  year={2015},
  organization={IEEE}
}

@article{li2020autonomous,
  title={Autonomous area exploration and mapping in underground mine environments by unmanned aerial vehicles},
  author={Li, Hang and Savkin, Andrey V and Vucetic, Branka},
  journal={Robotica},
  volume={38},
  number={3},
  pages={442--456},
  year={2020},
  publisher={Cambridge University Press}
}

@article{elmokadem2022method,
  title={A method for autonomous collision-free navigation of a quadrotor UAV in unknown tunnel-like environments},
  author={Elmokadem, Taha and Savkin, Andrey V},
  journal={Robotica},
  volume={40},
  number={4},
  pages={835--861},
  year={2022},
  publisher={Cambridge University Press}
}

@article{chung2023into,
  title={Into the Robotic Depths: Analysis and Insights from the DARPA Subterranean Challenge},
  author={Chung, Timothy H and Orekhov, Viktor and Maio, Angela},
  journal={Annual Review of Control, Robotics, and Autonomous Systems},
  volume={6},
  pages={477--502},
  year={2023},
  publisher={Annual Reviews}
}

@article{orekhov2022darpa,
  title={The DARPA subterranean challenge: A synopsis of the circuits stage},
  author={Orekhov, V and Chung, T},
  journal={Field Robotics},
  volume={2},
  number={1},
  pages={735--747},
  year={2022}
}

@article{cerberus2022,
author = {Marco Tranzatto  and Takahiro Miki  and Mihir Dharmadhikari  and Lukas Bernreiter  and Mihir Kulkarni  and Frank Mascarich  and Olov Andersson  and Shehryar Khattak  and Marco Hutter  and Roland Siegwart  and Kostas Alexis },
title = {CERBERUS in the DARPA Subterranean Challenge},
journal = {Science Robotics},
volume = {7},
number = {66},
pages = {eabp9742},
year = {2022},
doi = {10.1126/scirobotics.abp9742}}

@inproceedings{shi2019neural,
  title={Neural lander: Stable drone landing control using learned dynamics},
  author={Shi, Guanya and Shi, Xichen and O'Connell, Michael and Yu, Rose and Azizzadenesheli, Kamyar and Anandkumar, Animashree and Yue, Yisong and Chung, Soon-Jo},
  booktitle={2019 international conference on robotics and automation (icra)},
  pages={9784--9790},
  year={2019},
  organization={IEEE}
}

@article{cheeseman1955effect,
  title={The effect of the ground on a helicopter rotor in forward flight},
  author={Cheeseman, IC and Bennett, WE},
  year={1955}
}

@inproceedings{conyers2018empirical,
  title={An empirical evaluation of ceiling effect for small-scale rotorcraft},
  author={Conyers, Stephen A and Rutherford, Matthew J and Valavanis, Kimon P},
  booktitle={2018 International Conference on Unmanned Aircraft Systems (ICUAS)},
  pages={243--249},
  year={2018},
  organization={IEEE}
}

@article{prothin2019aerodynamics,
  title={Aerodynamics of MAV rotors in ground and corner effect},
  author={Prothin, S{\'e}bastien and Fernandez Escudero, Claudia and Dou{\'e}, Nicolas and Jardin, Thierry},
  journal={International Journal of Micro Air Vehicles},
  volume={11},
  pages={1756829319861596},
  year={2019},
  publisher={SAGE Publications Sage UK: London, England}
}

@article{tanabe2018multiple,
  title={Multiple rotors hovering near an upper or a side wall},
  author={Tanabe, Yasutada and Sugiura, Masahiko and Aoyama, Takashi and Sugawara, Hideaki and Sunada, Shigeru and Yonezawa, Koichi and Tokutake, Hiroshi},
  journal={Journal of Robotics and Mechatronics},
  volume={30},
  number={3},
  pages={344--353},
  year={2018},
  publisher={Fuji Technology Press Ltd.}
}

@article{nakata2020aerodynamic,
  title={Aerodynamic imaging by mosquitoes inspires a surface detector for autonomous flying vehicles},
  author={Nakata, Toshiyuki and Phillips, Nathan and Simoes, Patricio and Russell, Ian J and Cheney, Jorn A and Walker, Simon M and Bomphrey, Richard J},
  journal={Science},
  volume={368},
  number={6491},
  pages={634--637},
  year={2020},
  publisher={American Association for the Advancement of Science}
}

@misc{taffanel2021lighthouse,
      title={Lighthouse Positioning System: Dataset, Accuracy, and Precision for UAV Research}, 
      author={Arnaud Taffanel and Barbara Rousselot and Jonas Danielsson and Kimberly McGuire and Kristoffer Richardsson and Marcus Eliasson and Tobias Antonsson and Wolfgang Hönig},
      year={2021},
      eprint={2104.11523},
      archivePrefix={arXiv},
      primaryClass={cs.RO}
}

@inproceedings{osman2021tinyml,
  title={Tinyml platforms benchmarking},
  author={Osman, Anas and Abid, Usman and Gemma, Luca and Perotto, Matteo and Brunelli, Davide},
  booktitle={International Conference on Applications in Electronics Pervading Industry, Environment and Society},
  pages={139--148},
  year={2021},
  organization={Springer}
}

@inproceedings{de2021resilient,
  title={Resilient collision-tolerant navigation in confined environments},
  author={De Petris, Paolo and Nguyen, Huan and Kulkarni, Mihir and Mascarich, Frank and Alexis, Kostas},
  booktitle={2021 IEEE International Conference on Robotics and Automation (ICRA)},
  pages={2286--2292},
  year={2021},
  organization={IEEE}
}

@inproceedings{duisterhof2021sniffy,
  title={Sniffy bug: A fully autonomous swarm of gas-seeking nano quadcopters in cluttered environments},
  author={Duisterhof, Bardienus P and Li, Shushuai and Burgu{\'e}s, Javier and Reddi, Vijay Janapa and de Croon, Guido CHE},
  booktitle={2021 IEEE/RSJ International Conference on Intelligent Robots and Systems (IROS)},
  pages={9099--9106},
  year={2021},
  organization={IEEE}
}

@inproceedings{robinson2014computational,
  title={Computational investigation of micro rotorcraft near-wall hovering aerodynamics},
  author={Robinson, David Conal and Chung, Hoam and Ryan, Kris},
  booktitle={2014 international conference on unmanned aircraft systems (ICUAS)},
  pages={1055--1063},
  year={2014},
  organization={IEEE}
}

@inproceedings{hughes2021wall,
  title={Wall detection via IMU data classification in autonomous quadcopters},
  author={Hughes, Jason and Lyons, Damian},
  booktitle={2021 7th International Conference on Control, Automation and Robotics (ICCAR)},
  pages={189--195},
  year={2021},
  organization={IEEE}
}

@article{ding2023aerodynamic,
  title={Aerodynamic effect for collision-free reactive navigation of a small quadcopter},
  author={Ding, Runze and Bai, Songnan and Dong, Kaixu and Chirarattananon, Pakpong},
  journal={{NPJ} Robotics},
  volume={1},
  number={1},
  pages={2},
  year={2023},
  publisher={Nature Publishing Group UK London}
}

@inproceedings{abayan2022tof,
  title={ToF-based Simultaneous Localization and Mapping using a Shelled-UAV for Penstock Applications},
  author={Abayan, Jared Jan and Banglos, Charles Alver and Librado, Lester and Pao, Jeanette and Salaan, Carl John},
  booktitle={2022 IEEE 14th International Conference on Humanoid, Nanotechnology, Information Technology, Communication and Control, Environment, and Management (HNICEM)},
  pages={1--6},
  year={2022},
  organization={IEEE}
}

@article{ozaslan2017autonomous,
  title={Autonomous navigation and mapping for inspection of penstocks and tunnels with MAVs},
  author={{\"O}zaslan, Tolga and Loianno, Giuseppe and Keller, James and Taylor, Camillo J and Kumar, Vijay and Wozencraft, Jennifer M and Hood, Thomas},
  journal={IEEE Robotics and Automation Letters},
  volume={2},
  number={3},
  pages={1740--1747},
  year={2017},
  publisher={IEEE}
}

@article{niehorster2017accuracy,
  title={The accuracy and precision of position and orientation tracking in the HTC vive virtual reality system for scientific research},
  author={Niehorster, Diederick C and Li, Li and Lappe, Markus},
  journal={i-Perception},
  volume={8},
  number={3},
  pages={2041669517708205},
  year={2017},
  publisher={Sage Publications Sage UK: London, England}
}

@article{batterman1995hvac,
  title={HVAC systems as emission sources affecting indoor air quality: a critical review},
  author={Batterman, Stuart A and Burge, Harriet},
  journal={HVAC\&R Research},
  volume={1},
  number={1},
  pages={61--78},
  year={1995},
  publisher={Taylor \& Francis}
}

@phdthesis{vladthesis,
author = {Tempez, Vladislav},
year = {2022},
month = {06},
pages = {},
title = {Apprentissage d'une loi de commande optimale d'un petit quadrotor pour le vol dans des tuyaux cylindriques}
}

@book{aastrom2006advanced,
  title={Advanced PID control},
  author={{\AA}str{\"o}m, Karl Johan and H{\"a}gglund, Tore},
  year={2006},
  publisher={ISA-The Instrumentation, Systems and Automation Society}
}

@article{song2023reaching,
  title={Reaching the limit in autonomous racing: Optimal control versus reinforcement learning},
  author={Song, Yunlong and Romero, Angel and M{\"u}ller, Matthias and Koltun, Vladlen and Scaramuzza, Davide},
  journal={Science Robotics},
  volume={8},
  number={82},
  pages={eadg1462},
  year={2023},
  publisher={American Association for the Advancement of Science}
}

@article{macario2022comprehensive,
  title={A comprehensive survey of visual slam algorithms},
  author={Macario Barros, Andr{\'e}a and Michel, Maugan and Moline, Yoann and Corre, Gwenol{\'e} and Carrel, Fr{\'e}d{\'e}rick},
  journal={Robotics},
  volume={11},
  number={1},
  pages={24},
  year={2022},
  publisher={MDPI}
}

@article{rashad2020fully,
  title={Fully actuated multirotor UAVs: A literature review},
  author={Rashad, Ramy and Goerres, Jelmer and Aarts, Ronald and Engelen, Johan BC and Stramigioli, Stefano},
  journal={IEEE Robotics \& Automation Magazine},
  volume={27},
  number={3},
  pages={97--107},
  year={2020},
  publisher={IEEE}
}

@article{serres2017optic,
  title={Optic flow-based collision-free strategies: From insects to robots},
  author={Serres, Julien R and Ruffier, Franck},
  journal={Arthropod structure \& development},
  volume={46},
  number={5},
  pages={703--717},
  year={2017},
  publisher={Elsevier}
}

@article{castillo2023synchronization,
  title={Synchronization of a New Light-Flashing Shield with an External-Triggered Camera},
  author={Castillo-Zamora, Jose J and Negre, Amaury and Ingargiola, Jean-Marc and Ndoye, Abdoullah and Pouthier, Florian and Dumon, Jonathan and Durand, Sylvain and Marchand, Nicolas and Ruffier, Franck},
  journal={IEEE Sensors Letters},
  year={2023},
  publisher={IEEE}
}

@article{rak2007uhf,
  title={UHF propagation in caves and subterranean galleries},
  author={Rak, Milan and Pechac, Pavel},
  journal={IEEE Transactions on antennas and propagation},
  volume={55},
  number={4},
  pages={1134--1138},
  year={2007},
  publisher={IEEE}
}

@article{boutin2008radio,
  title={Radio wave characterization and modeling in underground mine tunnels},
  author={Boutin, Mathieu and Benzakour, Ahmed and Despins, Charles L and Affes, Sofi{\`e}ne},
  journal={IEEE Transactions on Antennas and Propagation},
  volume={56},
  number={2},
  pages={540--549},
  year={2008},
  publisher={IEEE}
}

@article{nikitin2003propagation,
  title={Propagation model for the HVAC duct as a communication channel},
  author={Nikitin, Pavel V and Stancil, Daniel D and Cepni, Ahmet G and Tonguz, Ozan K and Xhafa, Ariton E and Brodtkorb, Dagfin},
  journal={IEEE Transactions on Antennas and Propagation},
  volume={51},
  number={5},
  pages={945--951},
  year={2003},
  publisher={IEEE}
}

@article{bangar2021evaluating,
  title={Evaluating Performance of Heating, Ventilation \& Air Conditioning Duct Communication Channel at 60GHz Using Ray Tracing.},
  author={Bangar, Esha and Kiasaleh, Kamran},
  journal={Progress In Electromagnetics Research C},
  volume={113},
  year={2021}
}

@article{khalil2023autonomous,
  title={Autonomous Control of a Hybrid Rolling and Flying Caged Drone for Leak Detection in HVAC Ducts},
  author={Khalil, Ahmed and Jaradat, Mohammad A and Mukhopadhyay, Shayok and Abdel-Hafez, Mamoun F},
  journal={IEEE/ASME Transactions on Mechatronics},
  volume={29},
  number={1},
  pages={366--378},
  year={2023},
  publisher={IEEE}
}

@article{floreano2013miniature,
  title={Miniature curved artificial compound eyes},
  author={Floreano, Dario and Pericet-Camara, Ramon and Viollet, St{\'e}phane and Ruffier, Franck and Br{\"u}ckner, Andreas and Leitel, Robert and Buss, Wolfgang and Menouni, Mohsine and Expert, Fabien and Juston, Rapha{\"e}l and others},
  journal={Proceedings of the National Academy of Sciences},
  volume={110},
  number={23},
  pages={9267--9272},
  year={2013},
  publisher={National Acad Sciences}
}

@article{bergantin2023indoor,
  title={Indoor and outdoor in-flight odometry based solely on optic flows with oscillatory trajectories},
  author={Bergantin, Lucia and Coquet, C and Dumon, Jonathan and N{\`e}gre, Amaury and Raharijaona, Thibaut and Marchand, Nicolas and Ruffier, Franck},
  journal={International Journal of Micro Air Vehicles},
  volume={15},
  pages={17568293221148380},
  year={2023},
  publisher={SAGE Publications Sage UK: London, England}
}

@inproceedings{delgado2015vision,
  title={Vision-based humanoid robot navigation in a featureless environment},
  author={Delgado-Galvan, Julio and Navarro-Ramirez, Alberto and Nunez-Varela, Jose and Puente-Montejano, Cesar and Martinez-Perez, Francisco},
  booktitle={Pattern Recognition: 7th Mexican Conference, MCPR 2015, Mexico City, Mexico, June 24-27, 2015, Proceedings 7},
  pages={169--178},
  year={2015},
  organization={Springer}
}

@inproceedings{garrote2024exploiting,
  title={Exploiting 3D Grids for Indoor SLAM in Featureless Scenarios},
  author={Garrote, Luis and Reverendo, Ulisses and Nunes, Urbano J},
  booktitle={2024 IEEE International Conference on Autonomous Robot Systems and Competitions (ICARSC)},
  pages={151--156},
  year={2024},
  organization={IEEE}
}

@article{guzman2024tuning,
  title={Tuning rules for feedforward control from measurable disturbances combined with PID control: a review},
  author={Guzm{\'a}n, Jos{\'e} Luis and H{\"a}gglund, Tore},
  journal={International Journal of Control},
  volume={97},
  number={1},
  pages={2--15},
  year={2024},
  publisher={Taylor \& Francis}
}

@article{ebadi2023present,
  title={Present and future of slam in extreme environments: The darpa subt challenge},
  author={Ebadi, Kamak and Bernreiter, Lukas and Biggie, Harel and Catt, Gavin and Chang, Yun and Chatterjee, Arghya and Denniston, Christopher E and Desch{\^e}nes, Simon-Pierre and Harlow, Kyle and Khattak, Shehryar and others},
  journal={IEEE Transactions on Robotics},
  year={2023},
  publisher={IEEE}
}

@incollection{siciliano2016robotics,
  title={Robotics and the Handbook},
  author={Siciliano, Bruno and Khatib, Oussama},
  booktitle={Springer Handbook of Robotics},
  pages={1--6},
  year={2016},
  publisher={Springer}
}

@article{adrian1991particle,
  title={Particle-imaging techniques for experimental fluid mechanics},
  author={Adrian, Ronald J and others},
  journal={Annual review of fluid mechanics},
  volume={23},
  number={1},
  pages={261--304},
  year={1991}
}

@book{cavazzini2012particle,
  title={The Particle Image Velocimetry: Characteristics, Limits and Possible Applications},
  author={Cavazzini, Giovanna},
  year={2012},
  publisher={BoD--Books on Demand}
}

@article{wang2022neither,
  title={Neither fast nor slow: How to fly through narrow tunnels},
  author={Wang, Luqi and Xu, Hao and Zhang, Yichen and Shen, Shaojie},
  journal={IEEE Robotics and Automation Letters},
  volume={7},
  number={2},
  pages={5489--5496},
  year={2022},
  publisher={IEEE}
}

@inproceedings{wang2021estimation,
  title={Estimation and adaption of indoor ego airflow disturbance with application to quadrotor trajectory planning},
  author={Wang, Luqi and Zhou, Boyu and Liu, Chuhao and Shen, Shaojie},
  booktitle={2021 IEEE International Conference on Robotics and Automation (ICRA)},
  pages={384--390},
  year={2021},
  organization={IEEE}
}

@article{briod2014collision,
  title={A collision-resilient flying robot},
  author={Briod, Adrien and Kornatowski, Przemyslaw and Zufferey, Jean-Christophe and Floreano, Dario},
  journal={Journal of Field Robotics},
  volume={31},
  number={4},
  pages={496--509},
  year={2014},
  publisher={Wiley Online Library}
}

@inproceedings{martin2024compact,
  title={Compact docking station for sub-150g UAV indoor precise landing},
  author={Martin, Thomas and Blanco, Jefferson Roman and Mouret, Jean-Baptiste and Raharijaona, Thibaut},
  booktitle={2024 International Conference on Unmanned Aircraft Systems (ICUAS)},
  pages={1267--1274},
  year={2024},
  organization={IEEE}
}

@inproceedings{tuyishimire2017cooperative,
  title={Cooperative data muling from ground sensors to base stations using UAVs},
  author={Tuyishimire, Emmanuel and Bagula, Antoine and Rekhis, Slim and Boudriga, Noureddine},
  booktitle={2017 IEEE Symposium on Computers and Communications (ISCC)},
  pages={35--41},
  year={2017},
  organization={IEEE}
}


\section*{Supplementary Materials}
\begin{itemize}
\item Video: \url{https://youtu.be/BLQqoa7Zolw} -- The video shows the drone hovering in air ducts of different diameters.
\end{itemize}

\begin{si-figure*}
  \centering
  \includegraphics*[width=\linewidth]{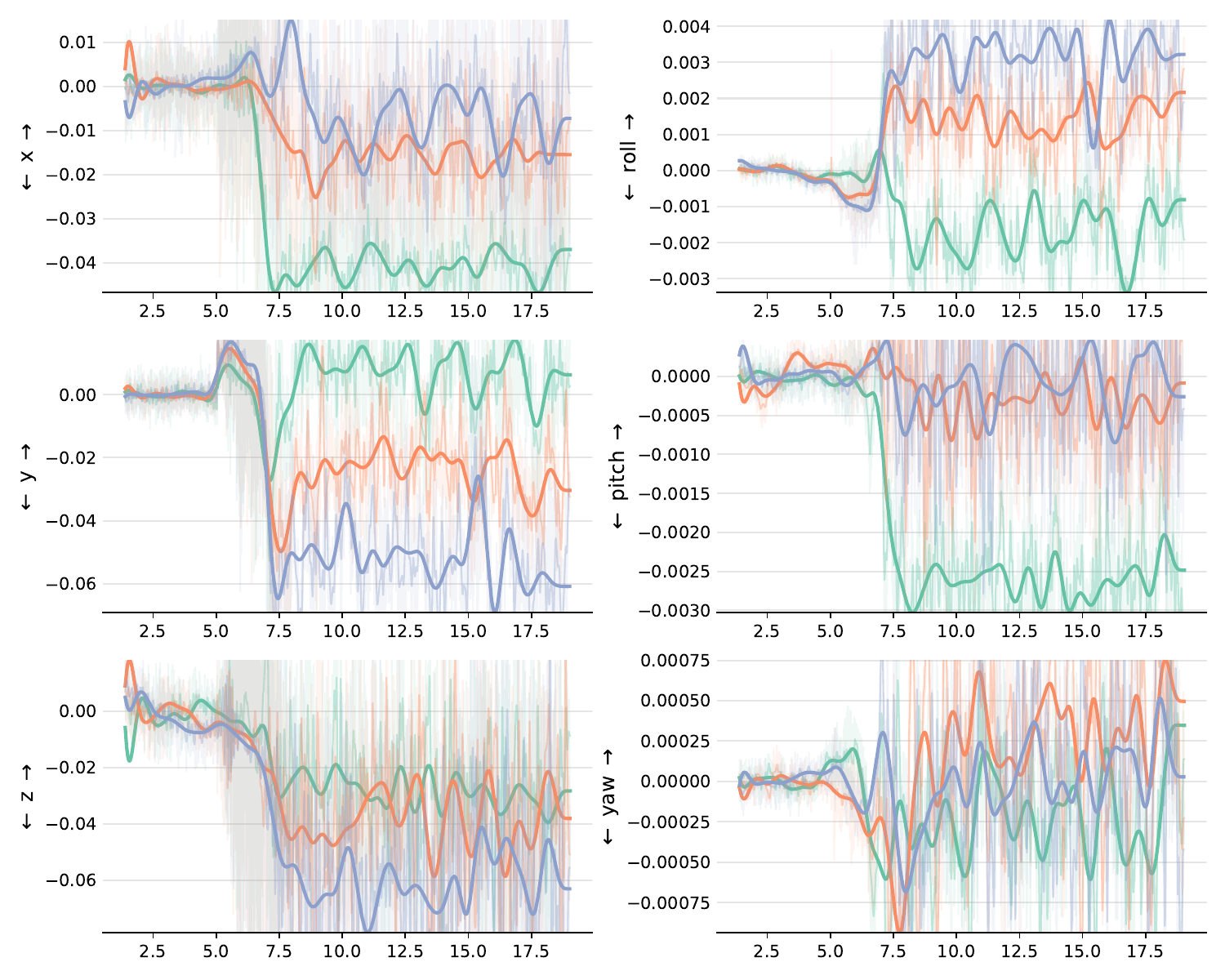}
  \caption{\label{si-fig:time}\textbf{Raw and filtered data from the force-torque sensor.} The three components of force and torque for a typical measure of additional force in an air duct, for three different positions (green, orange, and blue). The drone is first outside of the duct for 5 seconds, with little variations in force (in N) and torque (in N.m). Then, it enters the air duct and, for different positions, different forces appear. The data are filtered with a low-pass filter (Methods).}
\end{si-figure*}

\begin{si-figure*}
  \centering
  \includegraphics*[width=\linewidth]{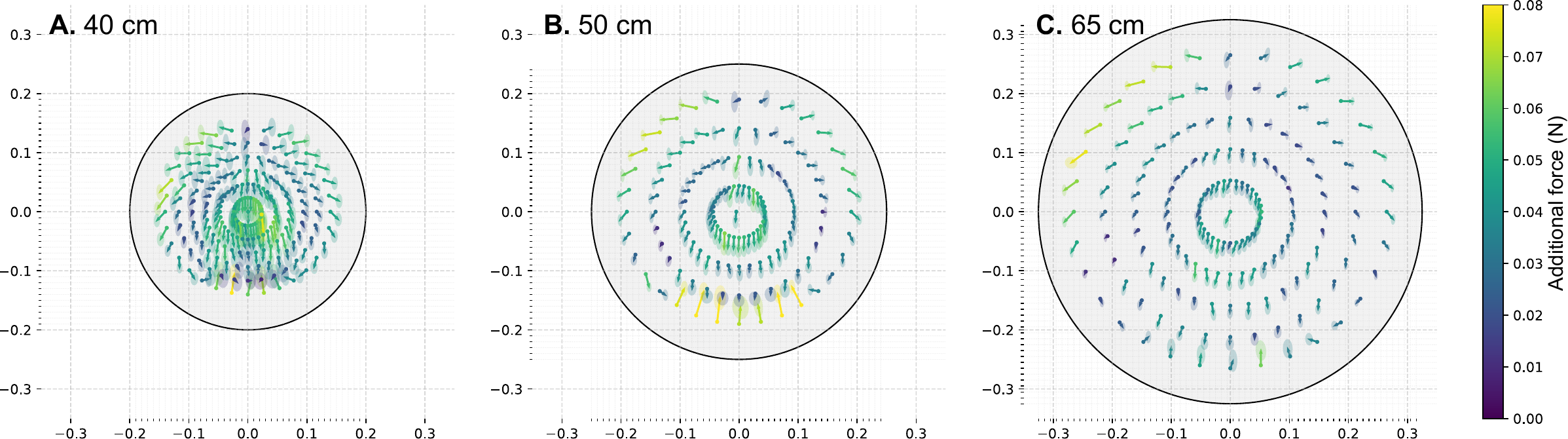}
  \caption{\label{si-fig:all-diameters}\textbf{Forces added by the duct for different diameters.} \textbf{A.} 40 cm. \textbf{B.} 50 cm. \textbf{C.} 65 cm. Overall, the force patterns stays similar and the force magnitude depends on the distance to the borders of the duct: lower forces in the middle of a larger duct than in a smaller one, but similar forces when close to the borders.}
\end{si-figure*}

\begin{si-figure*}
  \centering
  \includegraphics*[width=0.6\linewidth]{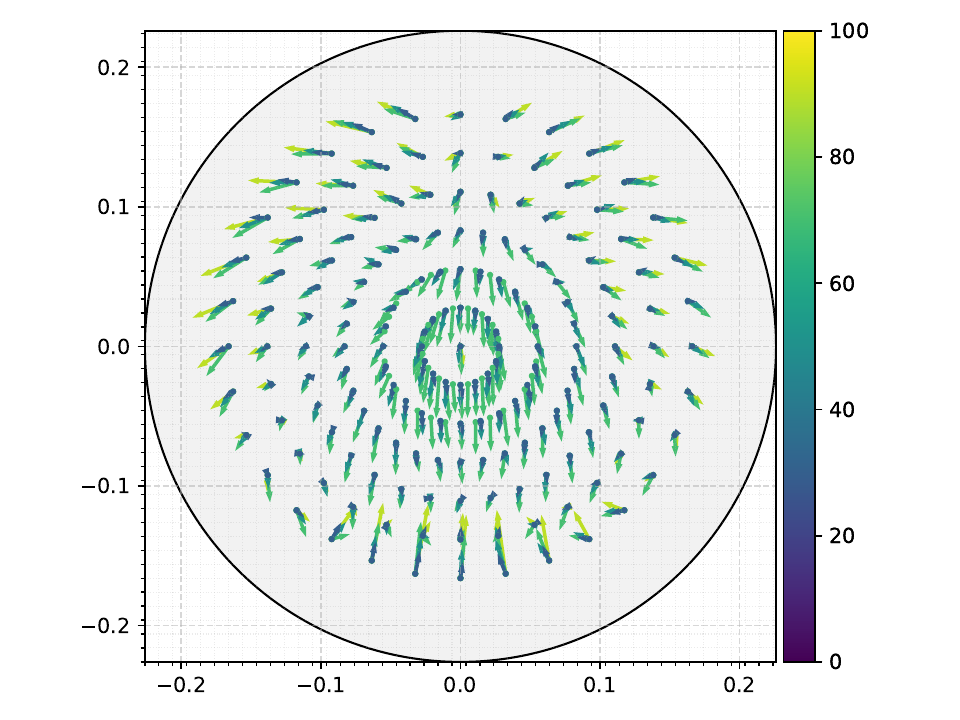}
  \caption{\label{si-fig:all-motors}\textbf{Influence of the motor regime.} Here, the color corresponds to the percentage of the max velocity of the motors. We tested 30\%, 50\% (about what is needed for hovering with our drone), 70\%, and 90\% in 45 cm circular air duct. Overall, the patterns stays the same but the intensity of the force increases with the motor regime (that is, the air velocity).}
\end{si-figure*}

\begin{si-figure*}
  \centering
  \includegraphics[width=\linewidth]{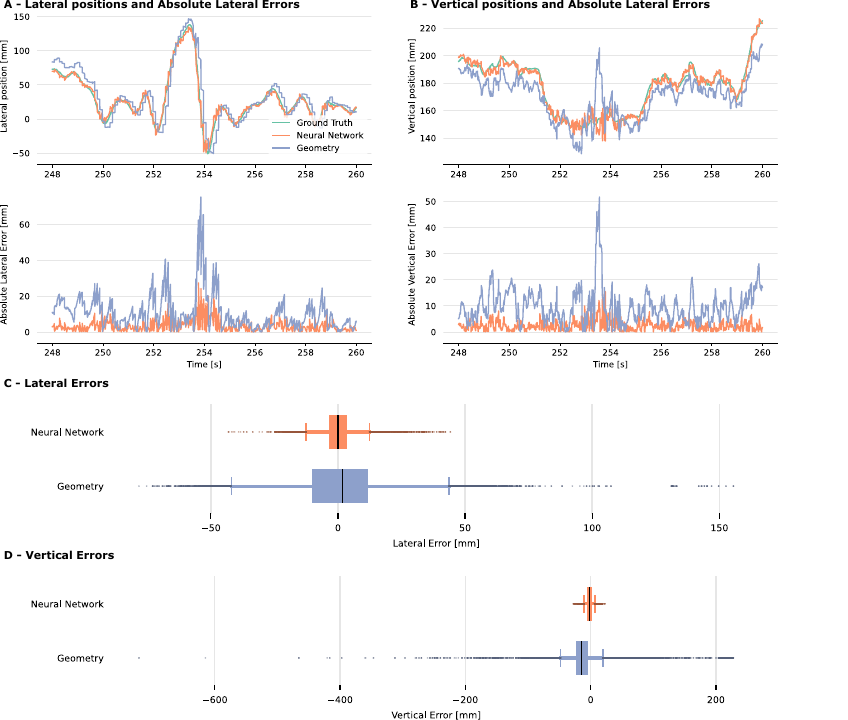}
  \caption{\label{si-fig:simul_results_45cm}
  \textbf{Comparison of the position estimations from the geometric and neural network methods against the ground truth inside a 45 cm air duct.} The green line represents the Ground Truth measured by the Extended Kalman Filter of the drone helped with the Lighthouse positioning system. The orange and blue lines represent respectively the lateral position outputted by the neural network and the geometrical solution.
  \textbf{A. Lateral positions and Absolute Lateral Errors.} In this time section of the test set, the geometrical solution is less accurate than the neural network.
  \textbf{B. Vertical positions and Absolute Vertical Errors.} In this time section of the test set, the geometrical solution is less accurate than the neural network.
  \textbf{C. Lateral Errors.} This boxplot represents the dispersion of the lateral position error outputted by the neural network and the geometrical solutions with respect to the ground truth in the test set. The neural network outputs a lateral position that is significantly more precise and accurate than the analytical approach, with a median error of -0.1 mm versus 1.8 mm laterally ([5\%,95\%] confidence interval [-8.4, 8.4] vs [-26.4, 29.7]).
  \textbf{D. Vertical Errors.} This boxplot represents the dispersion of the error between the vertical position measured by the neural network and the ground truth in the test set. The neural network outputs a vertical position that is significantly more precise and accurate than the analytical approach, with a median error of -1.8 mm versus -14.8 mm laterally ([5\%,95\%] confidence interval [-7.3, 4.3] vs [-42.9, 65.2]).}
\end{si-figure*}

\begin{si-figure*}
  \centering
  \includegraphics[width=\linewidth]{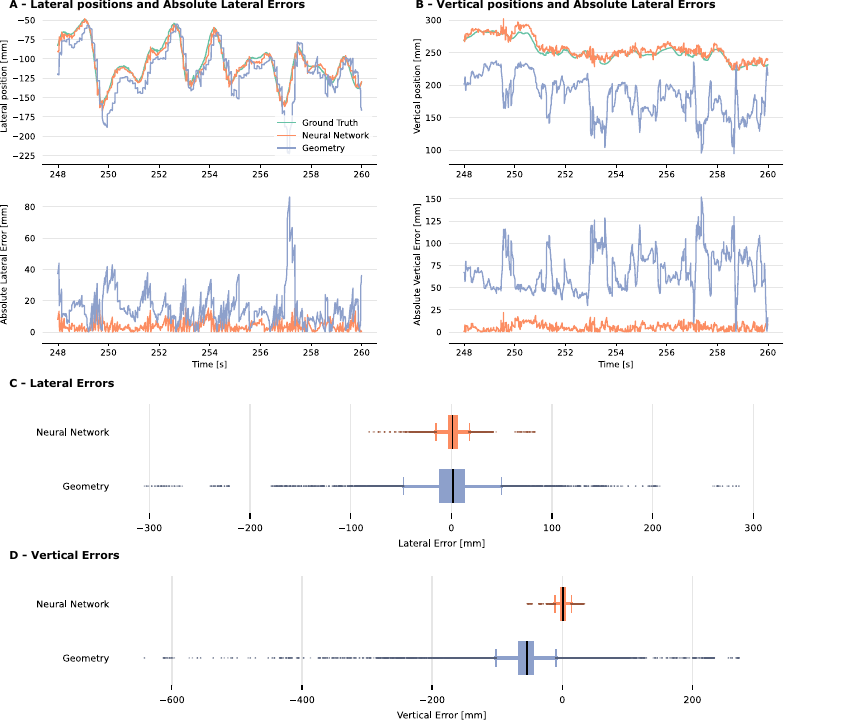}
  \caption{\label{si-fig:simul_results_56cm}
  \textbf{Comparison of the position estimations from the geometric and neural network methods against the ground truth inside a 56 cm air duct.} The green line represents the Ground Truth measured by the Extended Kalman Filter of the drone helped with the Lighthouse positioning system. The orange and blue lines represent respectively the lateral position outputted by the neural network and the geometrical solution.
  \textbf{A. Lateral positions and Absolute Lateral Errors.} In this time section of the test set, the geometrical solution is less accurate than the neural network.
  \textbf{B. Vertical positions and Absolute Vertical Errors.} In this time section of the test set, the geometrical solution is less accurate than the neural network.
  \textbf{C. Lateral Errors.} This boxplot represents the dispersion of the lateral position error outputted by the neural network and the geometrical solutions with respect to the ground truth in the test set. The neural network outputs a lateral position that is significantly more precise and accurate than the analytical approach, with a median error of 1.0 mm versus 1.7 mm laterally ([5\%,95\%] confidence interval [-10.3, 13.2] vs [-33.0, 30.5]).
  \textbf{D. Vertical Errors.} This boxplot represents the dispersion of the error between the vertical position measured by the neural network and the ground truth in the test set. The neural network outputs a vertical position that is significantly more precise and accurate than the analytical approach, with a median error of 1.2 mm versus -67.3 mm laterally ([5\%,95\%] confidence interval [-8.1, 11.2] vs [-98.5, -0.0]).}
\end{si-figure*}

\begin{si-figure*}
      \centering
      \includegraphics[width=0.72\linewidth]{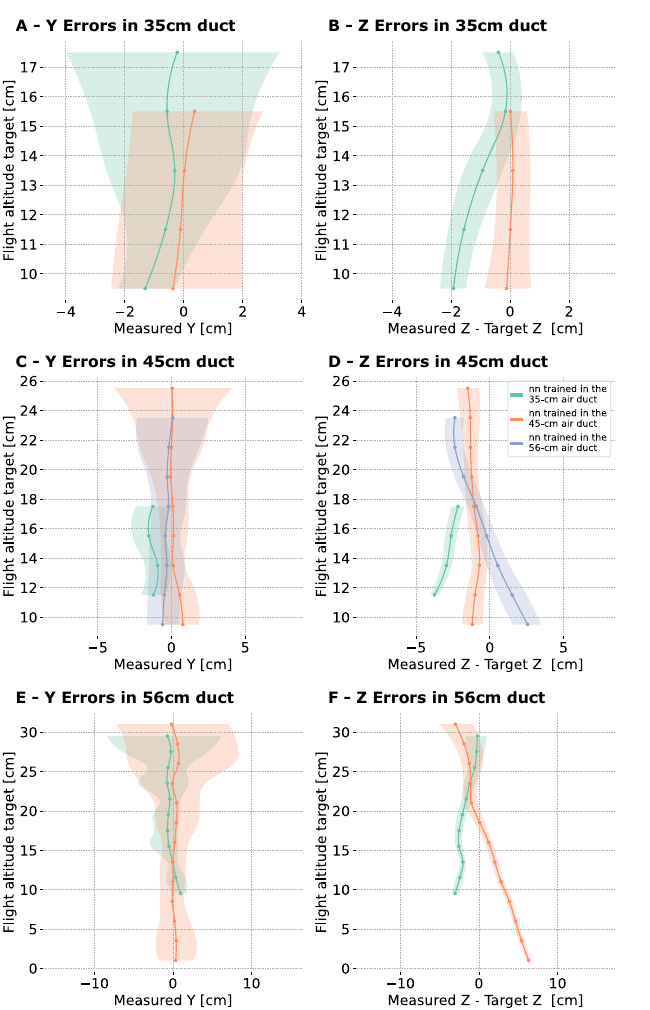}
      \caption{\label{si-fig:supplementary_stats}
      \textbf{Y and Z position errors for flights conducted in air ducts of 35, 45 and 56 cm in diameter using the Neural Networks (NN).} For each altitude target (depicted by the dots on the colored curves), 5 2-min flights are performed and the result are presented as the Interquartile Ranges (IQR) (in light color) and the medians (bright color lines with dots) of the lateral (Y) and vertical (Z) position errors. The NNs trained in the 35-cm, 45-cm and 56-cm air ducts are coded in green, orange and blue respectively. In general, we observe that the drone can fly in an air duct using an NN trained in one that is wider or narrower. However, in these situations, the capability to fly higher or lower decreases.
      \textbf{A-B. Y and Z Positions Errors in the 35-cm air duct.} In this air duct the drone can fly using the NNs trained in the 35-cm and 45-cm air ducts. The drone manages to fly in a cylinder 22\% narrower than the one where it has been trained on. However, the altitude range on which it is able to fly is very small, and under the middle of the air duct (17.5 cm). This could explain why the data show very slight increase in the IQR in the Y positions (Fig.~A) for the NN trained on the 45-cm air duct (orange), compared to the other one. On Fig.~B, the altitude precision of both datasets is comparable. The superior accuracy of the NN trained in the 45-cm air duct can be attributed to the higher quality of the dataset used for training.
      \textbf{C-D. Y and Z Positions Errors in the 45-cm air duct.} In this air duct the drone can fly using the NNs trained in the 35-cm, 45-cm and 56-cm air ducts. The drone manages to fly in a cylinder 29\% wider than the one where it has been trained on. Overall, Y errors (Fig.~C) depict a similar flight behaviour where the interquatile range gets greater as the altitude increases. The Z errors (Fig.~D) present a flight altitude that has a consistent accuracy using the NN trained in this air duct (nn 45 in orange). The other two present a stronger shift in accuracy.
      \textbf{E-F. Y and Z Positions Errors in the 56-cm air duct.} In this air duct the drone can fly using the NNs trained in the 45-cm and 56-cm air ducts. The Y errors (Fig.~E) depicts a similar interquartile range shape that widens around the middle of the air duct (22.5 cm). On Fig.~F, The NN trained in this airduct (blue) seems to output an altitude that shifts a lot or it could be the ground effect that pushes the drone up. On the same figure, the other NN (orange, NN 45) presents an underestimation of the altitude which corresponds to the observed IQR of NN 35 in green of Fig.~D, where both are flying in a wider airduct.}
\end{si-figure*}

\end{document}